\documentclass[12pt, oneside, a4paper]{report}
\usepackage{ifpdf}
\usepackage[bottom,hang]{footmisc}
\usepackage[a4paper, hdivide={3.2cm,,2cm}]{geometry}
\usepackage{acronym}
\usepackage[utf8]{inputenc} 
\usepackage[hyphens]{url}
\usepackage[dvips]{graphicx}
\usepackage[english]{babel} 
\usepackage{setspace}
\usepackage[fleqn]{amsmath}
\usepackage{wrapfig}
\usepackage{fancyhdr}
\usepackage[fit]{truncate}
\usepackage{courier}
\usepackage{caption}
\usepackage{needspace}
\usepackage[boxed,noline]{algorithm2e}

\usepackage{remreset}
\makeatletter
\@removefromreset{footnote}{chapter}
\makeatother

\usepackage{CJKutf8}
\usepackage[overlap, CJK]{ruby}
\usepackage{CJKulem}

\newenvironment{Japanese}{
\CJKfamily{min}
\CJKtilde
\CJKnospace}{}

\usepackage{enumitem}
\usepackage{listings} 

\LinesNumbered

\usepackage[numbers]{natbib}

\usepackage{color}

\definecolor{LinkColor}{rgb}{0,0,0.5}

\author{Paul Christian Sommerhoff}
\newcommand{\titleinfo}{Integration of Japanese Papers Into the DBLP Data Set}
\title{\titleinfo}

\definecolor{g}{gray}{0.4}

\begin{document}

\begin{titlepage}
\begin{center}
\begin{figure}
\begin{center}
  \includegraphics{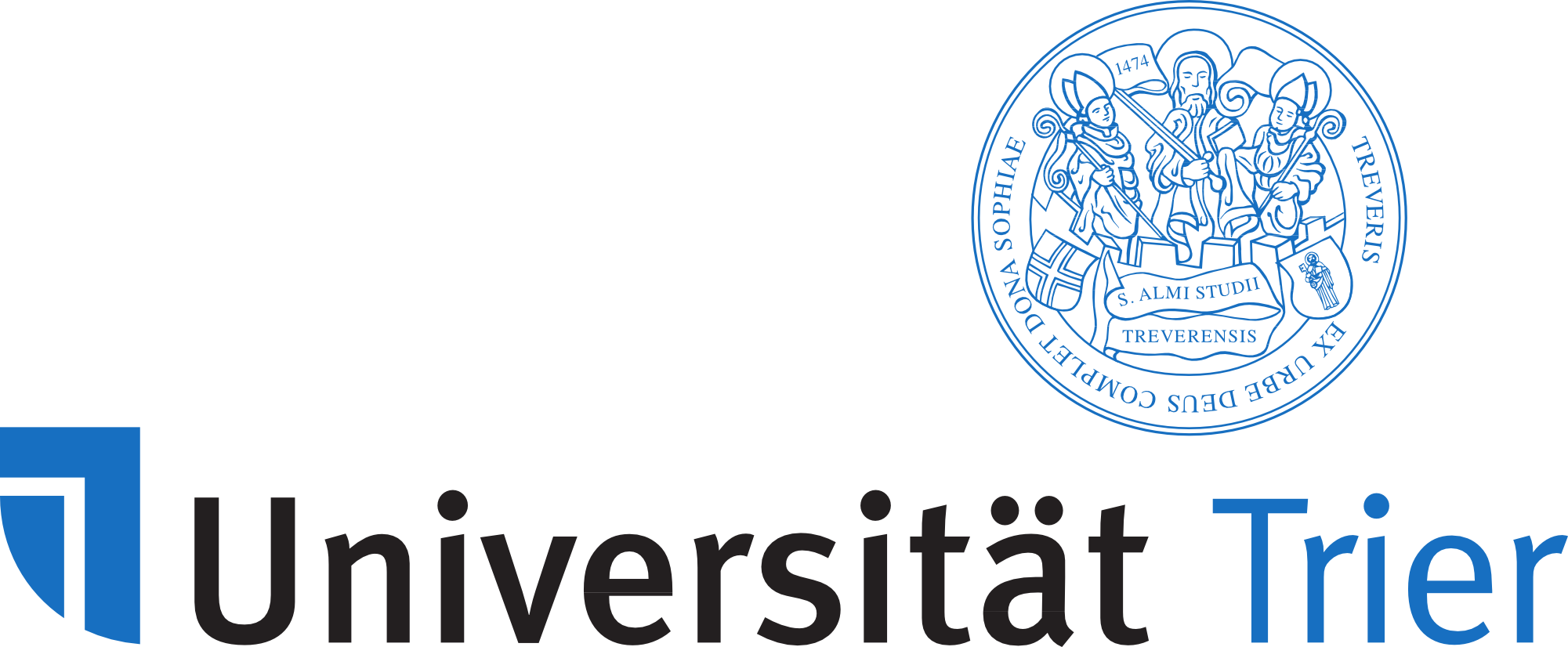}
\end{center}
\end{figure}
\vspace*{0 cm}
\vfill
\textsc{\LARGE Diplomarbeit}\\[0.4em]
zur Erlangung des akademischen Grades:\\
\textbf{Diplom-Informatiker}\\[0.4em]
\rule{\linewidth}{0.5mm} \\[0.4em]
\textbf{\huge Integration of Japanese Papers\\[0.2em]
Into the DBLP Data Set}
\rule{\linewidth}{0.5mm} \\[0.4em]
\vfill
\textit{Version for arXiv}\\
\begin{minipage}[t]{0.4\textwidth}
\begin{flushleft}
\emph{vorgelegt von}\\[0.5em]
Paul Christian Sommerhoff\\Matrikelnummer: XXXXXX\\[1em]
\emph{vorgelegt zum}\\[0.5em]
April 2013
\end{flushleft}
\end{minipage}
\begin{minipage}[t]{0.4\textwidth}
\begin{flushright}
\emph{vorgelegt am}\\[0.5em]
Lehrstuhl für\\Datenbanken und\\Informationssysteme\\der Abteilung
Informatik\\der Universität Trier\\[0.5em]
Prof. Dr. Bernd Walter
\end{flushright}
\end{minipage}
\end{center}
\end{titlepage}

\pagenumbering{Roman}
\bibliographystyle{is-alpha}

\newpage

\markright{\titleinfo}
\pagestyle{fancy}
\fancyhead{\protect\parbox[b]{\textwidth}{\textit{\leftmark}}}
\fancyfoot[C]{\thepage}

\onehalfspacing

\begin{abstract}
If someone is looking for a certain publication in the field of computer science, the searching person is likely to use the DBLP to find the desired publication.
The DBLP data set is continuously extended with new publications, or rather their metadata, for example the names of involved authors, the title and the publication date.
While the size of the data set is already remarkable, specific areas can still be improved. 
The DBLP offers a huge collection of English papers because most papers concerning computer science are published in English.
Nevertheless, there are official publications in other languages which are supposed to be added to the data set.
One kind of these are Japanese papers. 
\\This diploma thesis will show a way to automatically process publication lists of Japanese papers and to make them ready for an import into the DBLP data set.
Especially important are the problems along the way of processing, such as transcription handling and Personal Name Matching with Japanese names.
\end{abstract}
\newpage

\begin{CJK}{UTF8}{}
\begin{Japanese}

\tableofcontents
\markright{}
\newpage
\chapter*{List of Acronyms}
\addcontentsline{toc}{chapter}{List of Acronyms}
\markright{}
\begin{center}
\begin{acronym}[OAI-PMH]
 \acro{ACM}{Association for Computing Machinery}
 \acro{ASCII}{American Standard Code for Information Interchange}
 \acro{API}{Application Programming Interface}
 \acro{BHT}{Bibliography HyperText}
 \acro{DBLP}{Digital Bibliography \& Library Project \acroextra{(former meaning: DataBase systems and Logic Programming)}}
 \acro{FAQ}{Frequently Asked Questions}
 \acro{GB}{GigaByte}
 \acro{HTML}{HyperText Markup Language}
 \acro{HTTP}{HyperText Transfer Protocol}
 \acro{ID}{Identifier}
 \acro{IEEE}{Institute of Electrical and Electronics Engineers}
 \acro{IFIP}{International Federation for Information Processing}
 \acro{IPSJ}{Information Processing Society of Japan}
 \acro{IPSJ DL}{Digital Library of the Information Processing Society of Japan}
 \acro{ISO}{International Organization for Standardization}
 \acro{JAR}{Java ARchive}
 \acro{JDBC}{Java DataBase Connectivity}
 \acro{JDK}{Java Development Kit}
 \acro{OAI}{Open Archives Initiative}
 \acro{OAI-PMH}{Open Archives Initiative - Protocol for Metadata Harvesting}
 \acro{PDF}{Portable Document Format}
 \acro{RAM}{Random Access Memory}
 \acro{SAX}{Simple \acs{API} for \acs{XML}}
 \acro{SQL}{Structured Query Language}
 \acro{SPF}{Single Publication Format}
 \acro{TOC}{Tables Of Contents}
 \acro{URL}{Uniform Resource Locator}
 \acro{XML}{eXtensible Markup Language}
\end{acronym}
\end{center}
\newpage


\phantomsection
\addcontentsline{toc}{chapter}{List of Figures}
\listoffigures
\markright{}
\newpage

\pagenumbering{arabic}
\setcounter{page}{1}

\chapter{About This Diploma Thesis}\label{about}

The idea for this work was born when the author was searching for a possibility to combine computer science with his minor subject Japan studies in his diploma thesis. After dismissing some ideas leaning towards Named Entity Recognition and computer linguistics the author chose ``Integration of Japanese Papers Into the DBLP Data Set'' as his subject.
The \ac{DBLP} is a well-known and useful tool for finding papers published in the context of computer science.
The challenge to deal with such a huge database and the problems that occur when processing Japanese input data was the reason why this idea has been chosen. The hope is that, in the future, many Japanese papers can be added by the responsible people of the DBLP project.
\section{Motivation}\label{motivation}
Computer scientists are likely to use the DBLP to find information about certain papers or authors. 
Therefore, the DBLP is supposed to provide information about as many papers as possible.
For example, one could be interested in the paper ``Analysis of an Entry Term Set of a Civil Engineering Dictionary and Its Application to Information Retrieval Systems'' by Akiko Aizawa et al. (2005) but DBLP does not include it yet.
Japanese scientists might look for the original (Japanese) title ``土木関連用語辞典の見出し語の分析と検索システムにおける活用に関する考察'' or use Aizawa's name in Japanese characters (相澤彰子) for a search in DBLP.
The DBLP contains the author ``Akiko Aizawa'' but does not contain this specific paper or the author's original name in Japanese characters.
Our work is to implement a tool which addresses these questions, support the DBLP team in the integration of Japanese papers and reveal the difficulties of realizing the integration.
\section{Composition of the Diploma Thesis}\label{composition}
Dates are displayed in the \acs{ISO} 8601 standard format YYYY-MM-DD, e.g. 2012-10-19.
\\Although scientific works about the Japanese language often display the Sino-Japanese reading of \textit{kanji} (a Japanese character set) with uppercase letters to distinguish them from the other ``pure'' Japanese reading, we will not use uppercase letters to distinguish them in this work.
\\When a Japanese word is used in its plural form in this work, the word always stays unmodified. The reason is that in the Japanese language there is no differentiation between a singular and plural form.
\\We use a macron instead of a circumflex to display a long vowel of a Japanese word in Latin transcription (see section \ref{latin-chars}).
\section{Acknowledgement}\label{acknowledgement}
First I would like to thank Prof.~Dr.~Bernd Walter and Prof.~Dr.~Peter Sturm for making this diploma thesis possible.
Special thanks go to Florian Reitz for the great support and the useful answers for the questions I had while I have been working on this diploma thesis.
I also want to acknowledge the help of Peter Sommerhoff, Daniel Fett, David Christ and Kana Matsumoto for proofreading my work.
I thank Dr.~Michael Ley, Oliver Hoffmann, Peter Birke and the other members of the Chair of Database and Information Systems of the University of Trier.
Last but not least I want to tell some personal words to my family in my and their native language German:
\begin{quote}
 Ich möchte nun noch meinen Eltern und meinem Bruder Peter dafür danken, dass sie mich in meiner Diplomarbeitsphase, meinem Studium und auch schon davor immer unterstützt haben
 und immer für mich da waren, wenn ich sie brauchte. Ich weiß\ es zu schätzen.
\end{quote}
\chapter{Writing in Japanese}\label{writing-jap}
\begin{center}
``My view is that if your philosophy is not unsettled daily \\then you are blind to all the universe has to offer.'' \\(Neil deGrasse Tyson)
\end{center}

First we need to understand some aspects of the Japanese language and especially the different ways of writing Japanese because the peculiarities of the Japanese writing system are a crucial point of our work.
It lays the foundation for all Japanese-related subjects such as the structure of Japanese names (discussed in section \ref{structure-jnames}), a dictionary for Japanese names (discussed in section \ref{enamdict}) or the publication metadata source for Japanese publications (discussed in section \ref{ipsj-server}).
\\Hadamitzky (\cite{cite-hada}, p. 8-57) gives an overview about the basics of Japanese writing. The Japanese writing system includes \textit{kanji}, \textit{hiragana}, \textit{katakana} and the possibility to use Latin characters. 
\section{Kanji}\label{kanji}
\textit{Kanji} is the Japanese script which consists of traditional Chinese characters. It came to Japan around the 4th century. Since the Japanese had not developed an own writing system yet they began to use the Chinese characters. At the beginning, the characters were linked phonetically with a certain sound, so that they could write down all existing words by their sound. Applying this principle the \textit{man'y\=ogana} were created. Every character had one defined way to pronounce it. In addition to this, a second principle was introduced to write Japanese. This time the people orientated themselves on the meaning of the Chinese characters to choose a writing for a word. Applying the second principle, the \textit{kanji} were created. While the \textit{man'y\=ogana} were simplified to \textit{hiragana} and \textit{katakana} (see following sections \ref{hiragana} and \ref{katakana}) the general usage of \textit{kanji} did not change.
\\Due to an increase in number and possible readings of characters, the government began to try to simplify the Japanese writing system after the Meiji Restoration at the end of the 19th century. The last important reform took place after World War II. Along with some other changes and regulations, the permitted characters in official documents (\textit{t\=oy\=o kanji}) were limited to 1850 in 1946 and increased to 1900 in a draft from 1977. In 1981 they were replaced by the ``List of Characters for General Use'' (\textit{j\=oy\=o kanji}\footnote{for further information about \textit{j\=oy\=o kanji} (and modern Japanese writing) see \cite{cite-history}, p. 165-171}) containing 1945 characters. In 1951 the government published a list of additional 92 \textit{kanji} permitted for personal names. The number of \textit{kanji} permitted for personal names increased with time passing by. Eschbach-Szabo (\cite{cite-eschbach}, p. 175) says the last change permitted 983 \textit{kanji} for personal names in 2004. 
The press tries to abide by the \textit{j\=oy\=o kanji}. Japanese literature (science, fiction, etc.) uses about 4000 characters (comprehensive Sino-Japanese \textit{kanji} dictionaries contain ca. 10000 characters). 
Japanese people know approximately 3000 \textit{kanji} on average.
\\Due to their capability to give a word a meaning, \textit{kanji} are used in substantives, verbs, adjectives and Japanese personal names.
\\An important aspect is reading a \textit{kanji} because there are several possibilities to read one. Sait\=o and Silberstein (\cite{cite-saito}, p. 31-34) describe how to read a \textit{kanji}. There is a Japanese reading \textit{kun} and a Sino-Japanese reading \textit{on}. Depending on the text and grammar context either the \textit{kun} or \textit{on} reading is required. For example the \textit{kanji} 生\ is read \textit{sei} in 学生\ (\textit{gakusei}, meaning: student, \textit{on} reading) but is read $u$ in 生まれる\ (\textit{umareru}, meaning: being born, \textit{kun} reading). A single \textit{kanji} can have several \textit{kun} and several \textit{on} readings.
\\For our work it is important to know that one character can have several readings in names too.
\section{Hiragana}\label{hiragana}
The syllabary \textit{hiragana} evolved from the \textit{man'y\=ogana} by simplifying the characters. Every syllable is phonetically assigned to one sound of the spoken language (with two exceptions which can have two sounds each). The \textit{goj\=uon} table shown in figure \ref{pic:hiragana} lists the 46 syllables used today in a certain way (it can be compared with the ABC for letters).\footnote{Koop and Inada (\cite{cite-koop}, p. 20-29) provide information about \textit{hiragana}} Another but obsolete way to order the syllables is \textit{iroha} which is a poem containing all syllables. Although the name implies 50 sounds (\textit{goj\=u} means ``50'', \textit{on} means ``sound'') there are only 46 syllables left in modern Japanese. Actually, only 45 syllables belong to the \textit{goj\=uon} table. The $n$ counts as extra symbol (see \textit{goj\=uon} tables in figures \ref{pic:hiragana} and \ref{pic:katakana}).
\begin{figure}[bt]
\begin{center}
  \caption{Hiragana \textit{goj\=uon} table}\label{pic:hiragana}
  \includegraphics[scale=0.25]{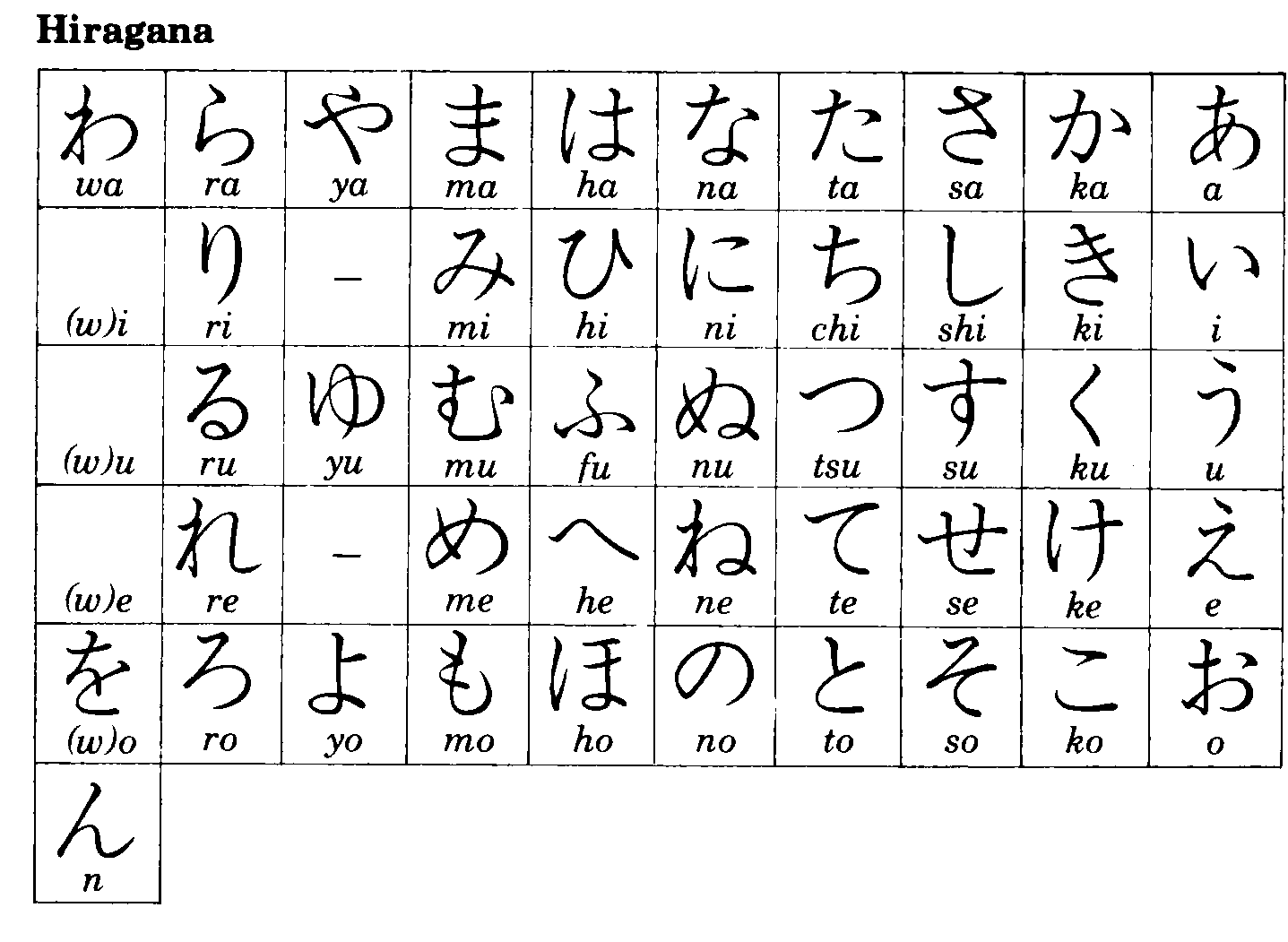}
  \caption*{Source: \cite{cite-saito}, p. 28}
\end{center}
\end{figure}
Other additional syllables are \textit{dakuon} (e.g. だ/$da$, recognizable by two little strokes), \textit{handakuon} (e.g. ぱ/$pa$, recognizable by a little circle) and \textit{y\=oon} (e.g. しゃ/$sha$, recognizable by a normally sized character that is followed by a smaller character).
\\You can write every Japanese word in \textit{hiragana} but if possible, \textit{kanji} are usually preferred to avoid problems with homonyms (we take a look at homonyms in chapter \ref{p-name-matching}). \textit{Hiragana} is mainly used to write words not covered by \textit{kanji} and as inflected endings. \textit{Kanji} and \textit{hiragana} are often combined within one word. For example 読む\ (\textit{yomu}) is the basic form of the verb ``to read''. The \textit{kanji} 読\ means reading by itself and in combination with the \textit{hiragana} syllable む\ it becomes the verb ``to read'' in a special grammatical form specifying tense, politeness level and other properties.\footnote{see also \cite{cite-genki} to get a beginner's guide to the Japanese language}
\section{Katakana}\label{katakana}
The syllabary \textit{katakana} also evolved from the \textit{man'y\=ogana} by simplifying the characters, consists of 46 characters nowadays (representing the same syllables as \textit{hiragana}) and is usually ordered by the \textit{goj\=uon} table. Figure \ref{pic:katakana} presents the \textit{katakana} in a \textit{goj\=uon} table. Besides optical differences with \textit{hiragana}, \textit{katakana} are used in other contexts. Japanese mostly use them to write foreign words including foreign personal names.
\begin{figure}[bt]
\begin{center}
  \caption{Katakana \textit{goj\=uon} table}\label{pic:katakana}
  \includegraphics[scale=0.25]{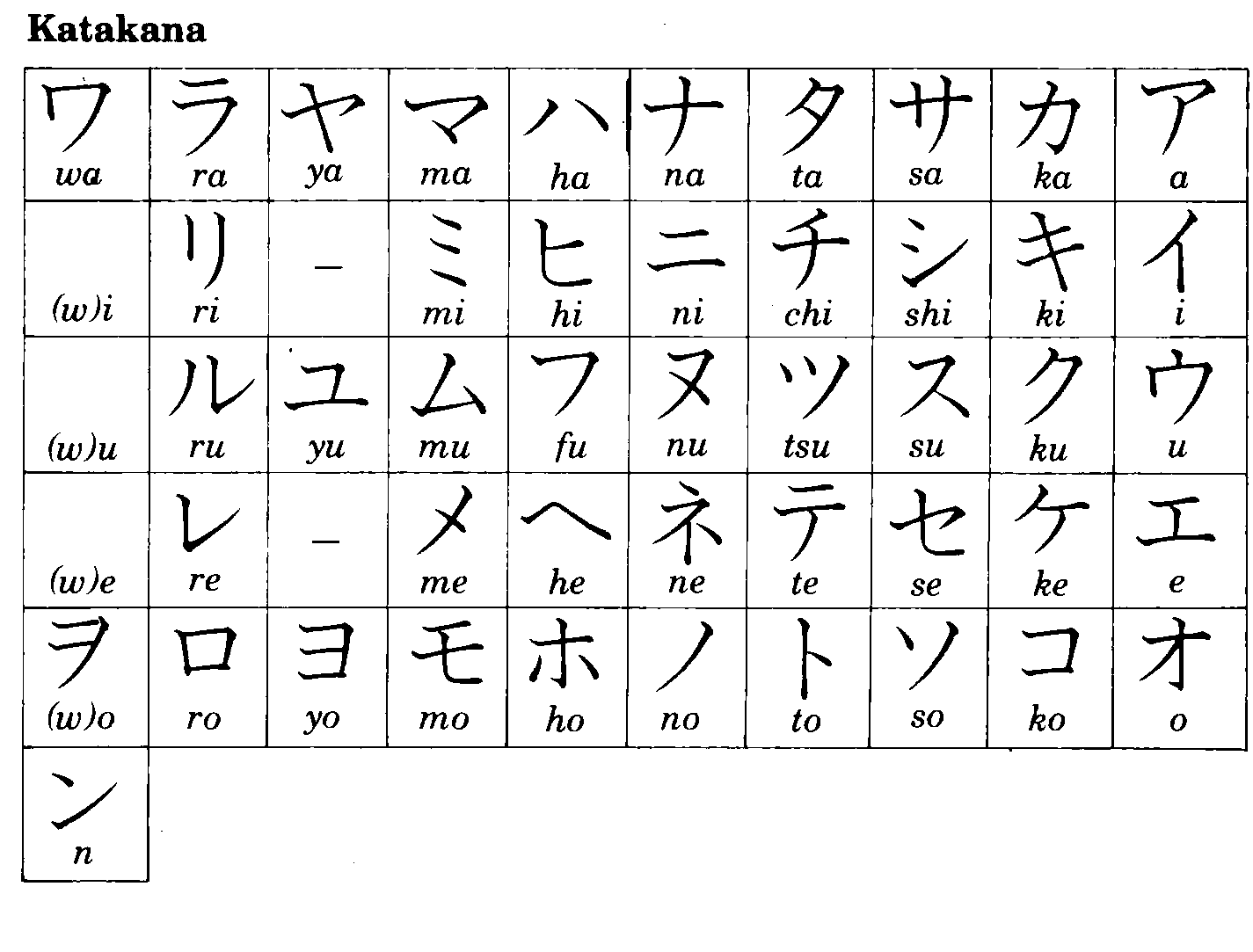}
  \caption*{Source: \cite{cite-saito}, p. 28}
\end{center}
\end{figure}
So foreigners often apply \textit{katakana} for their names. For example, the author's name can be transcribed as パウル·ソマホフ. The dot ·\ in the middle separates family and given name. Foreign names are often written with the given name preceding the family name.
\begin{figure}[ht]
\begin{center}
  \caption{``Hiragana'' written in \textit{hiragana}, ``katakana'' written in \textit{katakana}, ``kanji'' written in \textit{kanji}}\label{pic:kana-kanji}
  \includegraphics[scale=0.25]{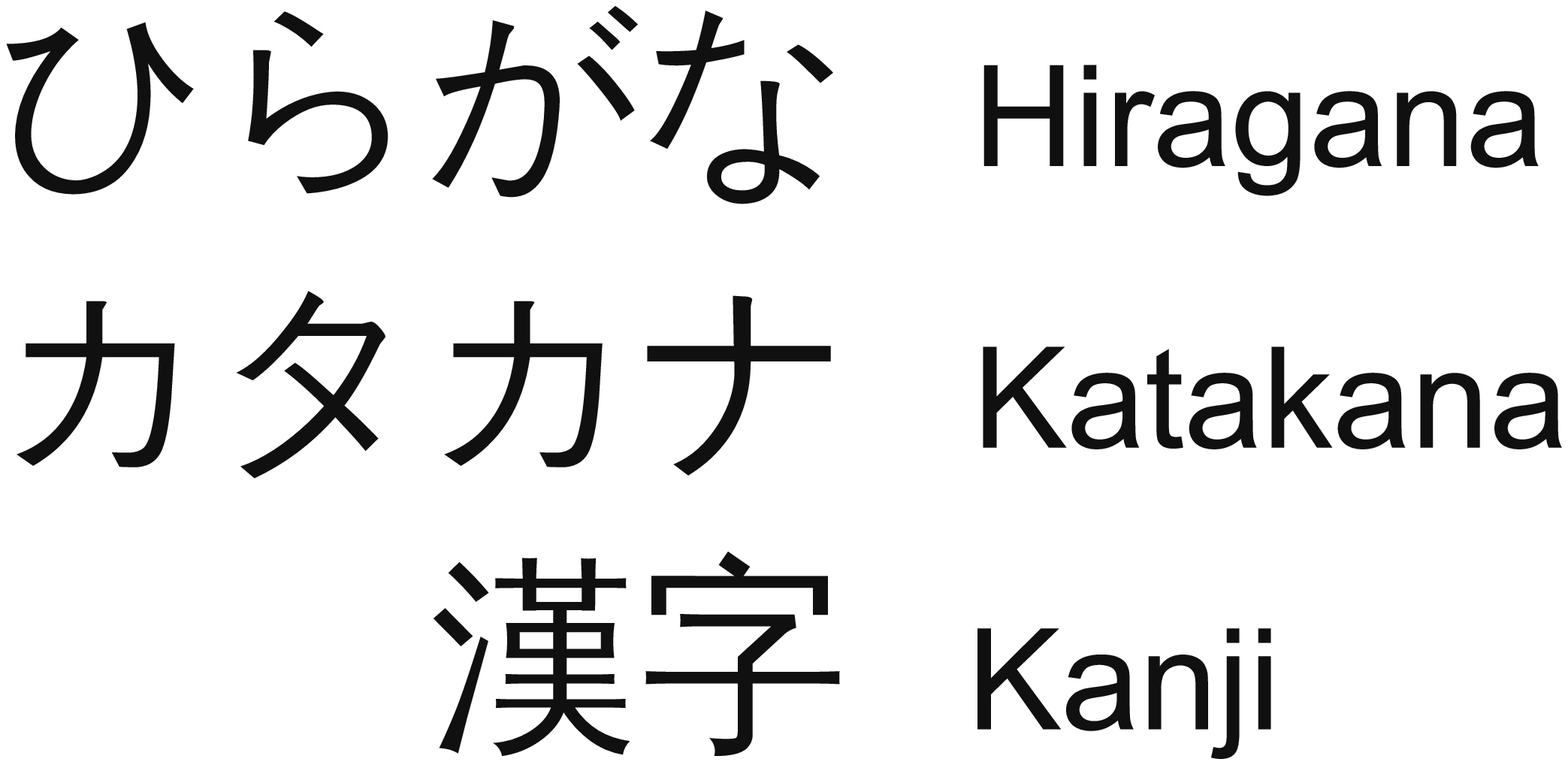}
  \caption*{Self-made creation}
\end{center}
\end{figure}
\section{Latin Characters/Transcription}\label{latin-chars}
Transcription systems which convert \textit{kanji}, \textit{hiragana} and \textit{katakana} to Latin characters are usually called \textit{r\=omaji}. Japanese can be easily transcribed by 22 letters and two additional signs. Due to many words having the same pronunciation, the meaning of words is sometimes ambiguous if they are transcribed into Latin characters. In 1954 the government released recommendations for transcribing Japanese. It recommended following two transcription systems:
\begin{description}
  \item[Kunreishiki r\=omaji] \hfill \\
  The \textit{kunreishiki\ r\=omaji} assigns transcriptions according to the order in the \textit{goj\=uon} table without regard to phonetic divergences of some consonants (we will discuss these divergences later). It has been introduced for official usage by the government only slightly different in 1937. It became the preferred transcription system in the standard \acs{ISO} 3602 ``Documentation - Romanization of Japanese (\textit{kana} script)'' \cite{cite-iso3602}.
  \item[Hebonshiki r\=omaji] \hfill \\
  The \textit{hebonshiki\ r\=omaji} was developed by a council of Japanese and foreign erudites in 1885 and spread by the American missionary  James C. Hepburn (\textit{Hebon} in Japanese), especially thanks to his Japanese-English dictionary published one year later. This work also employs \textit{hebonshiki}. \textit{Kunreishiki} would lead to transcriptions like \textit{kunreisiki}, \textit{hebonsiki} and \textit{kanzi}.
\end{description}
Although the \textit{kunreishiki} became the preferred system of the government, the international community often prefers the Hepburn system because the written words suggest a more intuitive pronunciation than \textit{kunreishiki}.
There are also language-related transcription systems that are rarely used. Kaneko and Stickel (\cite{cite-kaneko}, p. 53-55) mention them:
\begin{table}[ht]
\begin{center}
\begin{tabular}{| l | c | c | c | c |}
  \hline
  \textbf{Language} & \textbf{しゅ} & \textbf{し} & \textbf{つ} & \textbf{ち}\\
  \hline
  Portuguese (Jesuits) & xu & xi & t\c{c}u & chi\\
  \hline
  Dutch (\textit{rangakusha}) & sju & si & toe & ti\\
  \hline
  German (Siebold) & sju & si & tsu & tsi\\
  \hline
  French (Landresse) & siou & si & tsou & tsi\\
  \hline
  English (Hepburn) & shu & shi & tsu & chi\\
  \hline
\end{tabular}
\caption{Language-related transcription systems, source: \cite{cite-kaneko}, p. 53}
\label{tab:kaneko}
\end{center}
\end{table}
\\The important aspect are the system differences because we need to know where they occur when we deal with Personal Name Matching problems later. Figure \ref{pic:app-trans-diffs} in the appendix reveals the differences between the transcription systems. It summarizes 18 differences in all syllables including $dakuon$, $handakuon$ and $y\bar{o}on$. 
Unfortunately, there can be even more transcription differences.
\acs{ISO} 3602 highlights some more special cases when it comes to transcribing Japanese. One is the question whether to put an apostrophe after an $n$. To avoid misunderstandings, one should put an apostrophe behind an $n$ in certain cases. Otherwise, people could misinterpret the syllable $n$ followed by a syllable composed of a vowel or ``y'' and a vowel as syllables \textit{na}, \textit{ni}, \textit{nu}, \textit{ne}, \textit{no}, \textit{nya}, \textit{nyu} or \textit{nyo}. We will outline a practical example of this case in section \ref{impl-harvester}.
\\A second irregularity occurs when the same vowel appears right after another. If there is a morpheme boundary between the vowels, they should be transcribed as ``aa'', ``ii'', etc. but should be transcribed by an additional circumflex otherwise. 
\\Koop and Inada \cite{cite-koop} write about another difficulty called \textit{nigori}. 
\begin{quote}
``The \textit{nigori} (濁, literally, `turbidity', `impurity') ... [means] modifying the pronunciation of the consonant in certain of the kana sounds. It may be either (1) inherent, as in \textit{suge} (`sedge'), \textit{suzu} (`grelot'), \textit{go} (`five'), or (2) applied incidentally to the initial consonant of a word or name-element following another in composition, e.g., \textit{Shimabara} from \textit{shima} and \textit{hara}, \textit{nenj\=u} from \textit{nen} and \textit{ch\=u}, \textit{Harada} from \textit{hara} and \textit{ta}.'' (\cite{cite-koop}, p. 34) 
\end{quote}
So, if we want to derive a transcription from the family name 中田, we cannot tell whether to take \textit{Nakata} or \textit{Nakada} as the rightful transcription.
\chapter{Japanese Personal Names}\label{jnames}
\begin{center}
七転び、八起き。\quad Nana korobi, ya oki.
\\(Fall seven times, get up eight times.)
\\\textit{Japanese saying}
\end{center}

One of the central problems in this work is to deal with Japanese personal names. We need to get a picture of Japanese personal names in general to deal with multiple data sources (like the introduced publication metadata sources in chapter \ref{pub-metadata-sources}) which may represent the same name with different scripts or transcription methods. The dictionary ENAMDICT\footnote{will be introduced in section \ref{enamdict}} will be very helpful when it comes to extracting and verifying name information.
\section{Structure of Japanese Names}\label{structure-jnames}
Having the urge to name things is part of the human nature. Names make it easy to refer to things, people or any other object in this world. When it comes to name giving, history shows a development in the Japanese society. 
\\
Japanese names are divided into family and given name, similar to the system in the Western culture. When Japanese write their name in \textit{kanji} they put the family name first, followed by the given name (usually without leaving spaces between them), for example 中村武志\footnote{this name can also be found in the ENAMDICT dictionary as particular person born in 1967}\ (Takeshi Nakamura).\footnote{We intentionally use the expressions ``given name'' and ``family name'' instead of ``first name'' and ``last name'' in this work because of these circumstances and to avoid confusion.} While introducing themselves, they often tell their family name and skip the given name.
When Japanese refer to others, they have many name particles they put after a name to express the relationship to the other person. There is the neutral \textit{san}, \textit{chan} for children, \textit{kun} particular for boys or \textit{sensei} for teachers and doctors. (\cite{cite-genki}, p. 18-19)
\\\\Kagami (\cite{cite-namenforschung}, p. 913) writes about Japanese personal names. Only the samurai and nobility were allowed to carry family names before the Meiji Restoration in 1868. Merchants carried shop names instead (recognizable by the suffix \textit{-ya}), for example \textit{Kinokuniya} (shop name) \textit{Bunzaemon} (given name).
Then everybody had to pick a family name after the Meiji Restoration. 
Approximately 135000 family names are recognized now.
The most common family names are \textit{Suzuki}, \textit{Sat\=o}, \textit{Tanaka}, \textit{Yamamoto}, \textit{Watanabe}, \textit{Takahashi}, \textit{Kobayashi}, \textit{Nakamura}, \textit{It\=o}, \textit{Sait\=o} and others.
\begin{quote}
``In the feudal age, first and second given names were used as male names. The first name was \textit{Kemyoo} which was the order of brothers, and the second name was the formal name given at the coming of age ceremony (\textit{genpuku}), e.g. the name of a famous general in 12c.: \textit{Minamoto} (family name) \textit{no} (of) \textit{Kuroo} (kemyoo) \textit{Yoshitune} (formal given name), and before the \textit{genpuku} cere\-mony, he was called by \textit{Yoomyoo} (child name) \textit{Ushiwakamaru}.'' (\cite{cite-namenforschung}, p. 913)
\end{quote}
While there were no restrictions to the number of personal names visible until the Meiji Restoration, due to modernization, Japanese people got the restriction to carry only one given and one family name. (\cite{cite-eschbach}, p. 167-169)
\\Some indicators for assigning the gender to a name also exist. The suffixes \textit{-ko} (e.g. \textit{Hanako}), \textit{-mi} (\textit{Natsumi}) and \textit{-yo} (\textit{Yachiyo}) indicate a female name. Male names are harder to identify because they have no fixed pattern. The suffix \textit{-o} (\textit{Kazuo}) mostly belongs to a male name though.
\\Family names often consist of two \textit{kanji} characters, rarely of one or three characters. (\cite{cite-namenforschung}, p. 913)
\\\\Eschbach-Szabo (\cite{cite-eschbach}, p. 157-309) dedicates an elaborate chapter to Japanese personal names. Compared to the Chinese system, the Japanese naming system shows more tolerance. Several readings are left besides each other, formal rules are not always applied in practice. Japanese apprehend names mainly visually by the characters, secondarily by the reading and sound. This is why several readings for a written name are still acceptable in the modern Japanese world. 
In the feudal system, names were needed to determine the position and roles of a person in the family and the society rather than characterizing him or her as an individual.
Japan has an open naming system which allows adding new names. This is a difference to the exclusive name lists in Germany or France. (\cite{cite-eschbach}, p. 157-166)
\\Even the apparently simple \textit{kanji} 正\ has a lot of possible readings: \textit{Akira}, \textit{Kami}, \textit{Sada}, \textit{Taka}, \textit{Tadashi}, \textit{Tsura}, \textit{Nao}, \textit{Nobu}, \textit{Masa}. We can see the same phenomenon in recently approved \textit{kanji} too. When we see 昴\ we cannot be sure whether it is read \textit{K\=o} or \textit{Subaru}. (\cite{cite-ogawa})
\begin{quote}
``Conversely, it often happens that one does not know to write a name of given pronunciation. For example, Ogawa can be written 尾川\ or 小川. In Japan, when two people meet for the first time, they exchange business cards. This custom often baffles foreigners, but for Japanese it is a ritual with practical purpose: Japanese do not feel at ease until they see how a name is spelled out in kanji.'' (\cite{cite-ogawa})
\end{quote}
Figure \ref{pic:kobayashi151} illustrates the problem. The cashier tries to read the customer's name and cannot determine the right name. According to the customer's reaction, his first two trials \textit{Hiroko} and \textit{Y\=uko} seem to be wrong.
Ogawa considers the name polygraphy as a reason why the creation of new name characters is still allowed.
\begin{figure}[ht]
\begin{center}
  \caption{``Legibility is particularly important!!''}\label{pic:kobayashi151}
  \includegraphics[width=\textwidth]{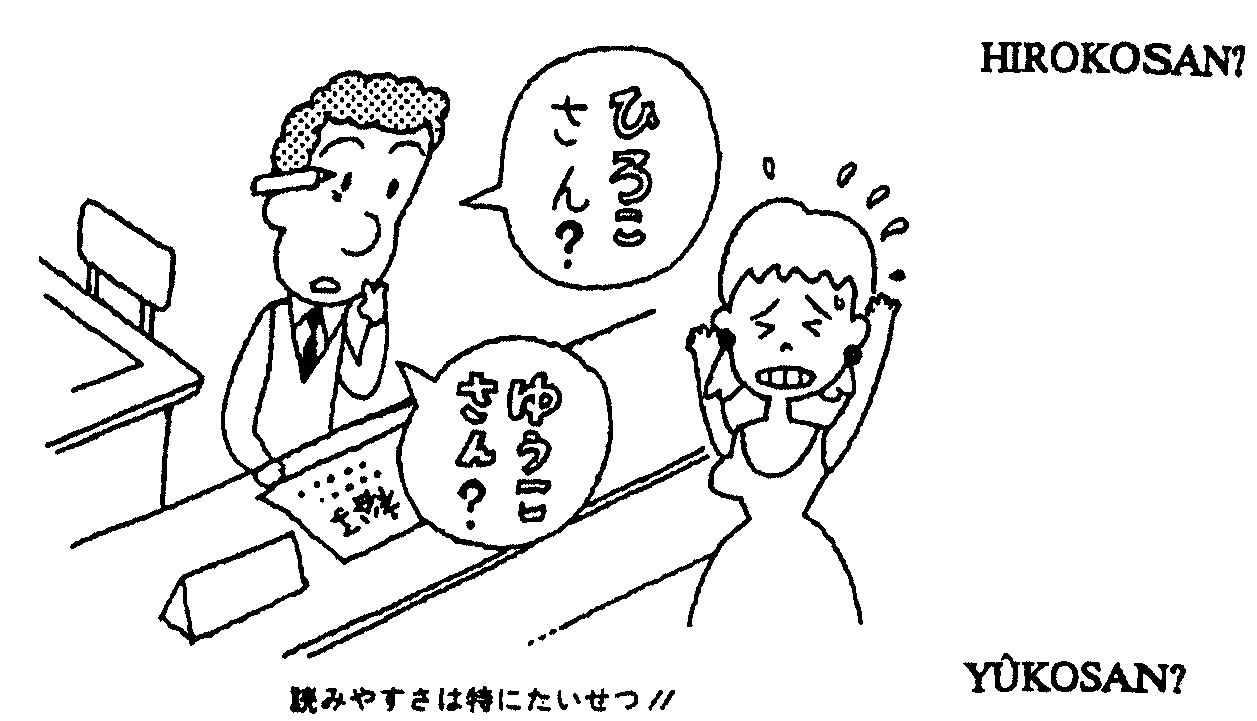}
  \caption*{Source: \cite{cite-kobayashi}, p. 151}
\end{center}
\end{figure}
\\
\begin{minipage}{\textwidth}
Some characteristics of the Japanese naming system are:
\begin{itemize}
  \item only little renaming of people
  \item semantic variance (names indicate different meanings/attributes)
  \item admission of foreign elements (foreign names get assimilated)
  \item possibility of polygraphic writing
  \item diversity of writing (many scripts usable, weak orthographic normalization)
  \item number of personal names for one person
\end{itemize}
\end{minipage} 
\vspace{0.5em}
\\
In academic circles a Sino-Japanese reading led to a more reputable name. So the famous linguist 上田万年\ from the Meiji era became known as \textit{Kazutoshi Ueda} AND \textit{Mannen Ueda} (\textit{Mannen} is the Sino-Japanese \textit{on} reading, \textit{Kazutoshi} is the Japanese \textit{kun} reading).
Modern guidebooks underline that maybe one has to take a loan word from another language to find the corresponding reading for a name in \textit{kanji}. For example, 宇宙\ could be read as \textit{Kosumo} (from the Greek word for cosmos) instead of \textit{Uch\=u}. Also ノイ\ (\textit{Noi}), derived from the German word ``neu'' (new), became a Japanese given name. Another imaginable name is ``Sky'' written as 空海\ (meanings: 空\ Sky, 海\ sea) and transcribed as \textit{Sukai} (actually \textit{k\=ukai}). This would finally show the impact of globalization also on the Japanese naming system.
If one has lived in Japan for a while and wants to adapt or register his or her Western name, one can choose corresponding \textit{kanji} either by meaning or reading of the original name. Another possibility is transcribing the name with \textit{katakana}.  (\cite{cite-eschbach}, p. 170-171, 305-309)
\\The name Anna exists in many cultures. The girls in figure \ref{pic:eschbach309} are both called Anna. Both turn around when they hear their name and respond in their mother tongue (``Yes!'' and ``Hai!'', respectively).
\begin{figure}[ht]
\begin{center}
  \caption{Anna! Yes! Hai!}\label{pic:eschbach309}
  \includegraphics[width=0.5\textwidth]{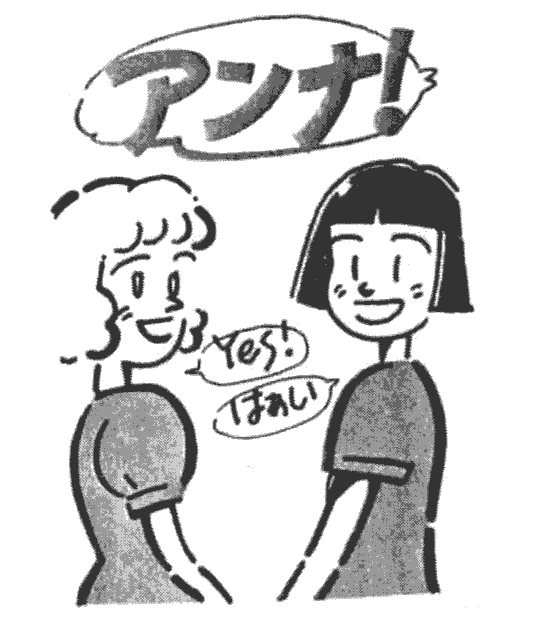}
  \caption*{Source: \cite{cite-eschbach}, p. 309}
\end{center}
\end{figure}
\\One principle of Japanese name giving is \textit{ateji}. \textit{Ateji} (当て字) means ``appropriate characters''. It says Japanese try to find characters with good, positive meanings for their children's name. Examples are 愛子\ (愛: \textit{ai}, love; 子: \textit{ko}, child), 夏美\ (夏: \textit{natsu}, summer; 美: \textit{mi}, beauty) or 正\ (\textit{Tadashi}, correct, honest). There is also a list with characters that are allowed but should be avoided because of bad associations. Characters like 蟻\ (\textit{ari}, ant), 苺\ (\textit{ichigo}, strawberry), 陰\ (\textit{kage}, shadow), 悪\ (\textit{aku}, bad/evil) belong to this list. (\cite{cite-eschbach}, p. 172-176)
\\A particular case drew public attention from June 1993 to February 1994 when \textit{Shigeru Sat\=o} wanted to call his son \textit{Akuma}, written as 悪魔\ (devil/demon). The civil registry office declined the registration after some discussion because they were worried about other children teasing him. The father went to court but the judges also declined the wish. Although the father wanted to give his son a unique, rememberable name, the judges saw a possible problem in his individual identification process and also getting teased (\textit{ijime}) by other children in school someday.
Then \textit{Sat\=o} tried to choose other characters while keeping the reading \textit{Akuma}. But also changing the name partly into \textit{man'y\=ogana} (亜久魔\footnote{亜: \textit{a}, asia; 久: \textit{ku}, long; 魔: \textit{ma}, ghost}) did not change anything about the declination because of the phonological equality implying the same negative associations.
Thereupon the father picked the character 神\ (god) and its unusual reading \textit{Jin}. Even though Shintoistic gods can be good or evil, the civil registry office accepted the name.
\textit{Sat\=o} announced his intention to keep calling his son \textit{Akuma} anyway. So a new (yet unofficial) reading for a character might be established. (\cite{cite-eschbach}, p. 271-278)
\\An article of ``Japan Today'' from December 2012 shows that there is still a debate about this subject.
\begin{quote}
``[...]Shinzo Abe\footnote{Shinzo Abe is currently Prime Minister of Japan (since 2012-12-26)}, the leader of the Liberal Democratic Party made a stand against kirakira\footnote{literally ``sparkling'', Japanese expression for flashy names} names last week when he stated that giving a child a name like Pikachu\footnote{Pikachu is one of the most famous Pok\'emon (Pok\'emon is a video game franchise owned by Nintendo)}, which could be written something like 光宙\ (`light' and `space'), is tantamount to child abuse, saying: `Children are not pets; we have to provide guidance for parents who would name their child in such a way.' ''(\cite{cite-pikachu})
\end{quote}
Despite regulations, the discussion about the culture of name giving does not seem to have ended yet.
Japanese comics like the one in figure \ref{pic:kobayashi155} suggest a happy-go-lucky life if one has a common everyday name like \textit{Keiko}.
\begin{figure}[ht]
\begin{center}
  \caption{``With an everyday name you can step through life steadily''}\label{pic:kobayashi155}
  \includegraphics[width=0.5\textwidth]{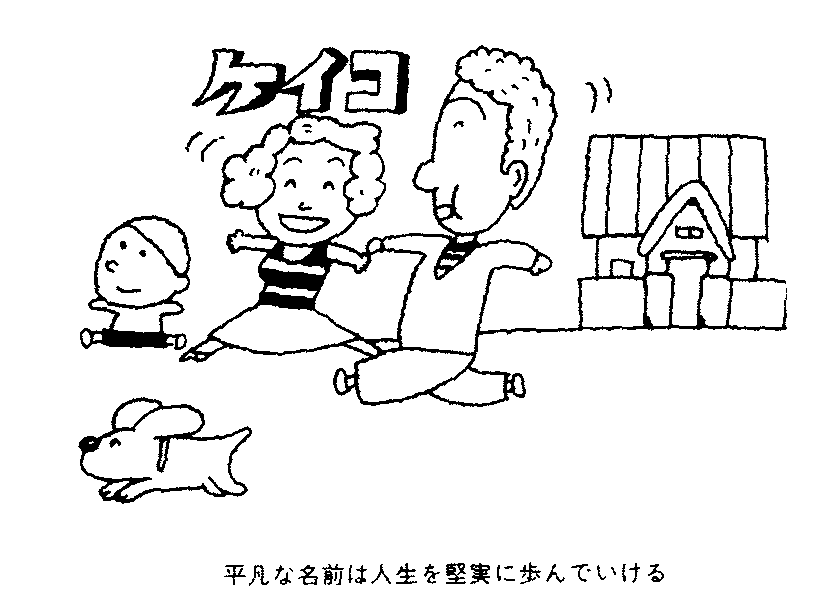}
  \caption*{Source: \cite{cite-kobayashi}, p. 155}
\end{center}
\end{figure}
\\Today's registration of names allows 2983 \textit{kanji} for given names, 4000 \textit{kanji} for family names, 700 \textit{man'y\=ogana}, 46 \textit{hiragana} and 46 \textit{katakana}. There are still people whose names are written with the obsolete kana syllabary \textit{hentaigana} which has been prohibited in 1948 (\cite{cite-eschbach}, p. 176-177; \cite{cite-hentaigana}). Regarding this variety of characters (and readings) it is not surprising that even well educated Japanese have problems reading certain names too, respectively they cannot be sure that the chosen reading is the correct reading in the current situation.
Forbidden is the usage of geometrical and punctuation signs. The sign ◯\ (\textit{maru}) is an example of such a forbidden one. Also forbidden is the usage of Latin characters (\textit{r\=omaji}) at the registration of a name. \textit{R\=omaji} can be used privately, though. (\cite{cite-eschbach}, p. 176-177)
\\Names can be changed by marriage, adoption or getting a pseudonym or special posthumous name. Titles can be acquired too. (\cite{cite-eschbach}, p. 251)
\\After disestablishing the patriarchal \textit{ie} system in which a man (for example the husband) is the dominating householder of a family, the family name has not been focused on the affiliation to a family anymore but has been focused on the couple living together in joint lives. (\cite{cite-eschbach}, p. 253-255)
\\\\Writing a Japanese name can be ambiguous. While the name written in \textit{kanji} is definite, displaying it in Latin characters leads to several possibilities. Japanese themselves usually write their name using \textit{kanji}. To find matching authors in the DBLP\footnote{we will discuss the DBLP project more detailed in section \ref{dblp}}, it will be crucial for us to have names in Latin characters later on (in chapter \ref{import}) because the standard encoding format of the file containing the main data of the DBLP project is ISO 8859-1 (Latin-1).
\\\\We sometimes talk about ``\textit{kanji} names'' or ``names in \textit{kanji} representation'' in this work. Although the expression does not suggest it, they shall include all names in Japanese characters, ergo names in \textit{kanji}, \textit{hiragana} and \textit{katakana}.
\section{ENAMDICT}\label{enamdict}
To automatically detect where a Japanese family name in \textit{kanji} notation ends and the given name begins, we should factor a name dictionary into our work. It is important that this dictionary includes the names written in \textit{kanji} and a clear transcription for them in Latin characters. A useful dictionary for our purposes is ENAMDICT.
\\ENAMDICT \cite{cite-enamdict} is a free dictionary for Japanese proper names, maintained by the Monash University in Victoria (Australia). The Electronic Dictionary Research and Development Group\footnote{\url{http://www.edrdg.org/}} owns the copyright. In 1995, ENAMDICT became an independent project by dividing the universal dictionary EDICT into two projects. ENAMDICT contains person names and non-person names like places and companies as well. Table \ref{tab:enamdict} shows the online statistics about the content of the ENAMDICT file. We will call the categories ``name types'' in subsequent chapters.
\begin{quote}
 ``A proper name is a word or group of words which is recognized as having identification as its specific purpose, and which achieves, or tends to achieve that purpose by means of its distinctive sound alone, without regard to any meaning possessed by that sound from the start, or aquired by it through association with the object thereby identified.'' (\cite{cite-gardiner}, p. 73)
\end{quote}
\begin{table}[ht]
\begin{center}
\begin{tabular}{| l | r | c |}
  \hline
  \textbf{Name category} & \textbf{Number of entries} & \textbf{abbreviation\footnotemark} \\
  \hline
  surnames & 138500 & s \\
  \hline
  place names & 99500 & p \\
  \hline
  unclassified names & 139000 & u \\
  \hline
  given names & 64600 & g \\
  \hline
  female given names & 106300 & f \\
  \hline
  male given names & 14500 & m \\
  \hline
  full names of particular persons & 30500 & h \\
  \hline
  product names & 55 & pr \\
  \hline
  company names & 34 & co \\
  \hline
  stations & 8254 & st \\
  \hline
\end{tabular}
\caption{Statistics about the ENAMDICT file, data source: \cite{cite-enamdict}, self-made creation}
\label{tab:enamdict}
\end{center}
\end{table}
\footnotetext{these intern abbreviations occur again when we construct a database for Japanese names in chapter \ref{enamdict-to-db}}
\chapter{Publication Metadata Sources}\label{pub-metadata-sources}
\begin{center}
百語より一笑 \quad Hyaku go yori issh\=o
\\(A smile is more worth than a hundred words.)
\\\textit{Japanese saying}
\end{center}

This chapter gives an overview of the publication metadata sources that we will need later. We take a look at these sources because we will discuss a way to extract metadata information from one source containing Japanese papers and import them into another source in chapter \ref{import}.
\section{Digital Library of the IPSJ}\label{ipsj-server}
The \ac{IPSJ} is a Japanese society in the area of information processing and computer science. It was founded in April 1960 and, by its own account, helps evolving computer science and technology and contributes new ideas in the digital age. It regularly publishes the magazine ``Information Processing'' (\textit{j\=oh\=o shori}) and a journal, holds symposiums and seminars, Special Interest Groups issue technical reports and hold conferences.
It is also the Japan representative member of the \ac{IFIP} and established partnerships with the \ac{IEEE}, \ac{ACM} and other organizations.
\enlargethispage{-2\baselineskip}
IPSJ develops drafts of international standards and Japanese industrial standards as well.
Eight regional research sections are widespread over Japan. IPSJ had over 17000 members in March 2011. (\cite{cite-ipsj-act}; \cite{cite-ipsj-org})
\\
The IPSJ provides a Digital Library\footnote{\url{https://ipsj.ixsq.nii.ac.jp/ej/}, accessed at 2012-10-10} (referenced as \acs{IPSJ DL} in this work) where everybody can search Japanese papers in the field of computer science.
The search page can be displayed in Japanese and English, most papers are written in Japanese. Free papers are accessible in \acs{PDF} format, non-free can be bought.
A tree view provides the order structure of the papers and there is a keyword search available.
We are especially interested in the metadata export functions, though. The online application offers following export formats:
\begin{itemize}
  \item \acs{OAI-PMH}
  \item BibTeX
  \item OWL SWRC
  \item WEKO Export
\end{itemize}
For our purposes the \ac{OAI-PMH} is the most suitable solution because we can send simple \acs{HTTP} requests to the server and get publication metadata as a result.
It ``provides an application-independent interoperability framework based on metadata harvesting'' (\cite{cite-oai-pmh}) and consists of two groups of participants. Data Providers can be servers hosting and supplying the metadata. Service Providers take the harvester role and process the recieved metadata from the Data Provider.
The application-independent interoperability is achieved by using \acs{XML} as basic exchange format.\footnote{further information about XML can be found in \cite{cite-beg-xml}} Arbitrary programs can parse XML input data very easily, so can we.
\begin{figure}[bt]
\begin{center}
  \caption{Part of publication metadata in OAI-PMH Dublin Core format}\label{pic:example-oai-pmh-dc}
  \includegraphics[scale=0.5]{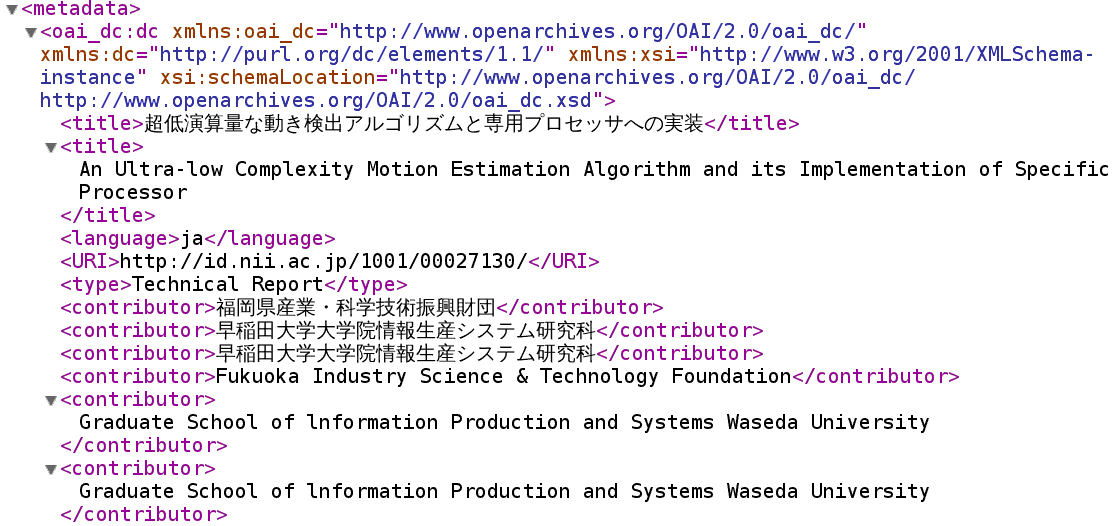}
\end{center}
\end{figure}
While accessing the server, the data can be extracted in several ways. We can either access an OAI-PMH repository by the repository name, the metadata format prefix of the record and a unique identifier\footnote{Figure \ref{pic:example-oai-pmh-dc} shows a part of the result of such a request.} or get a list of records with only one request.
\\
\begin{minipage}{\textwidth}
A request for a list of records looks like this:
\begin{lstlisting}[]{http}
http: //ipsj.ixsq.nii.ac.jp/ej/
  ?action=repository_oaipmh&verb=ListRecords
  &metadataPrefix=oai_dc
\end{lstlisting}
\end{minipage}
It may also contain a start date and an end date or a resumption token. The headers of records include a corresponding time stamp. The server's response to a request offers only 100 publications. We need this resumption token because it determines the point where we resume the harvest.
\\In the beginning and for debugging, it was more comfortable to increment a counter that acts as the unique identifier and send requests for single entries with the respective ID multiple times. Fortunately, the entries can be addressed by such an integer ID (plus some constant name):
\begin{lstlisting}[]{http2}
http: //ipsj.ixsq.nii.ac.jp/ej/
  ?action=repository_oaipmh&verb=GetRecord&metadataPrefix=oai_dc
  &(*@\textbf{identifier}@*)=oai:ipsj.ixsq.nii.ac.jp:(*@\textbf{27130} @*)
\end{lstlisting}
The last entry containing real publication metadata has the suffix integer 87045 in its ID.\footnote{The maximum ID was determined at 2012-11-13. Due to new publications the value is presumably higher in the future.} After that some entries with status $deleted$ follow. If we continue requesting even higher IDs, we soon get only a reply with the error code $idDoesNotExist$ anymore, implying there are no publications with higher IDs.
We will discuss the implementation of an OAI-PMH harvester for the IPSJ DL in section \ref{impl-harvester}.
\section{DBLP Project}\label{dblp}
The \ac{DBLP}\footnote{\url{http://dblp.uni-trier.de}, accessed at 2012-10-10} is a worldwide known database for publication metadata in the field of computer science.
Ley \cite{cite-ley-dblp} gives a brief explanation of the DBLP, additional information is extracted from the online DBLP \ac{FAQ} \cite{cite-dblp-faq}. It was started in 1993 as a test server for web technologies and named ``Database systems and Logic Programming'' in the beginning.
But it grew and became a popular web application for computer scientists. The Computer Science department of the University of Trier founded the project, since summer 2011 it is a joint project of Schloss Dagstuhl - Leibniz Center for Informatics and the University of Trier. 
\begin{quote}
``For computer science researchers the DBLP web site is a popular tool to trace the work of colleagues and to retrieve bibliographic details when composing the lists of references for new papers. Ranking and profiling of persons, institutions, journals, or conferences is another sometimes controversial usage of DBLP.'' (\cite{cite-ley-dblp})
\end{quote}
The publication metadata is stored in the \acs{XML} file $dblp.xml$\footnote{available at \url{http://dblp.uni-trier.de/xml/}} containing more than 2 million publications and exceeding a size of 1 GB (state of October 2012). An excerpt of the beginning of $dblp.xml$ can be found in the appendix section \ref{app-dblp.xml}. 
The header dictates ISO-8859-1 (Latin-1) as encoding of the file. Considering that we want to import Japanese names in \textit{kanji} (which are not included in Latin-1) we must handle that issue somehow. We will discuss the solution in section \ref{create-bht}.
\begin{figure}[!ht]
\begin{center}
  \caption{New logo of the DBLP project}\label{pic:dblp-logo}
  \includegraphics[width=0.7\textwidth]{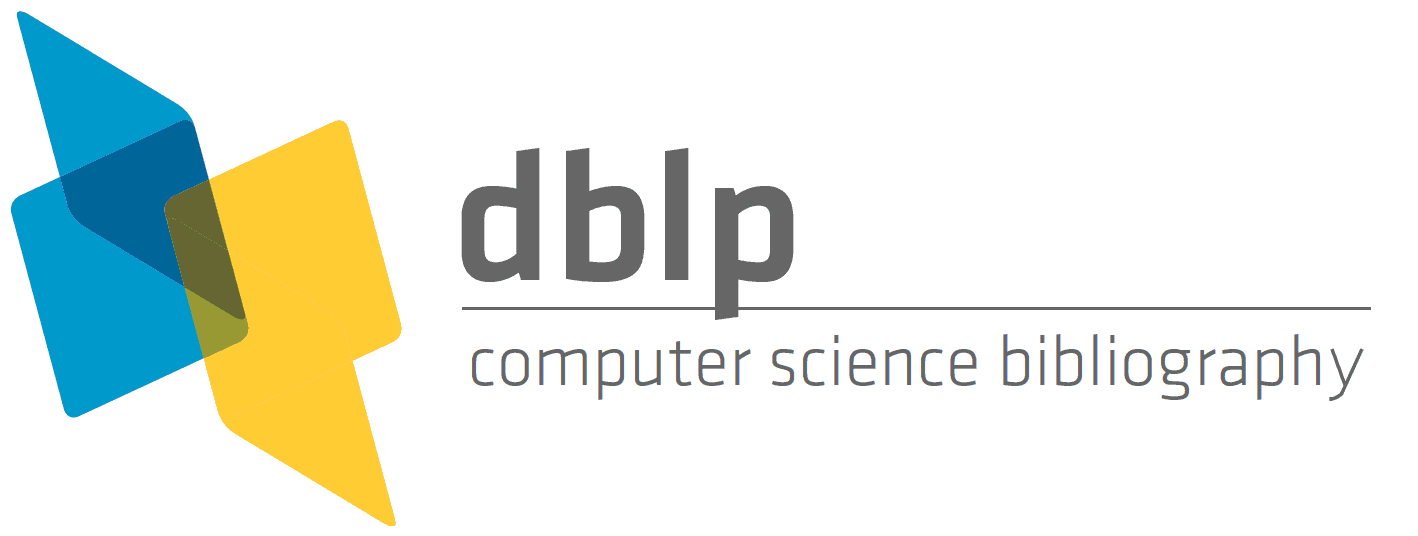}
  \caption*{Source: \cite{cite-dblp}}
\end{center}
\end{figure}
\\The web front end of the DBLP provides an overview of coauthor relationships by a Coauthor Index (see figure \ref{pic:coauthor-index}). The Coauthor Index can be found at the author's page after the list of the author's publications itself.
It shows all coauthors, common papers and categorizes the coauthors into groups that worked together by giving the author names corresponding background colors.
\begin{figure}[!ht]
\begin{center}
  \caption{Screenshot of the DBLP Coauthor Index of Atsuyuki Morishima}\label{pic:coauthor-index}
  \includegraphics[width=0.7\textwidth]{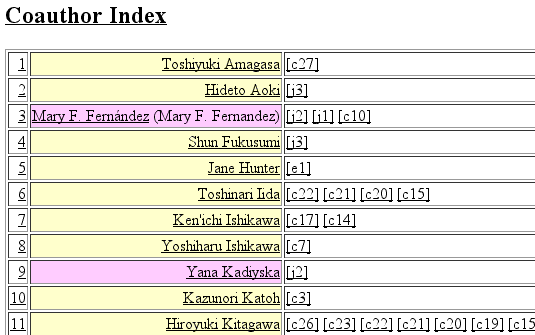}
  \caption*{Full Coauthor Index at \cite{cite-dblp-morishima}}
\end{center}
\end{figure}
\\In his diploma thesis Vollmer \cite{cite-vollmer} gives useful hints in terms of converting the $dblp.xml$ file to a relational database. He also compares the performance of several relational database management systems for this conversion.
\\\\The DBLP team developed a special format for the integration of new publications. It is called Bibliography Hypertext (\acs{BHT}), is based on \acs{HTML} and similar to the HTML code of the tables of contents (\acs{TOC}s) at the DBLP website.
An example of a publication list in BHT format can be found in the appendix in section \ref{app-dblp-bht}.
A BHT file has the following structure. The header (text between \textit{h2} tags) contains the volume, the number/issue and the date of issue.
A list of corresponding publications follows next. The list is surrounded by a beginning and a closing $ul$ tag, single publication entries start with a $li$ tag. A comma is used for the separation of authors while there should be a colon after the last author name. Then comes the title which has to end with a period, question mark or exclamation point.
The next line provides the start and end page in the volume/issue.
At last, an optional URL can be added by an $ee$ element to specify an ``electronic edition'' for a paper.
Some guidelines need to be considered, too:
\begin{itemize}
 \item there is no closing $li$ tag
 \item initials should be avoided (full name is preferred)
 \item titles with only upper case letters should be avoided
 \item ``0-'' is the default page number value if the page information is missing
\end{itemize}
The BHT file may contain additional information. For example, conference proceedings may have more headers to achieve a better clarity.
But it should be as close to the proposed format as possible to guarantee an easy import without unnecessary burdens.
(\cite{cite-hoffmann}; \cite{cite-dblp-faq}, ``What is the preferred format to enter publications into DBLP?'')
\\We will extend the original format in section \ref{create-bht} to satisfy our needs in the context of Japanese papers.
\chapter{Personal Name Matching}\label{p-name-matching}
\begin{center}
``The important thing is not to stop questioning; \\curiosity has its own reason for existing.'' \\(Albert Einstein)
\end{center}

After looking at transcription systems, Japanese personal names and publication metadata sources, we will now have to look at Personal Name Matching to enable us to deal with the Japanese names extracted from the metadata sources.
First we will discuss Personal Name Matching in general and then problems of Personal Name Matching for Japanese names in particular.
\\The expression \textit{Personal Name Matching} comes from the work by Borgman and Siegfried \cite{cite-borgman} and is used here as in the extended definition from Reuther's work (\cite{cite-reuther}, p. 48-51).
Borgman and Siegfried only talk about synonyms. Synonyms are possible names for the same person. Reuther extended the definition by also including homonyms. A name is a homonym if it can belong to several persons.
Personal Name Matching is known by other titles in literature, too. Niu et al. \cite{cite-niu} discuss \textit{Cross Document Name Disambiguation}:
\begin{quote}
``Cross document name disambiguation is
required for various tasks of know\-ledge discovery
from textual documents, such as entity tracking,
link discovery, information fusion and event
tracking. This task is part of the co-reference task:
if two mentions of the same name refer to same
(different) entities, by definition, they should
(should not) be co-referenced. As far as names are
concerned, co-reference consists of two sub-tasks:
\begin{enumerate}[label=(\roman*)]
\item name disambiguation to handle the problem of different entities happening to use the same name;
\item alias association to handle the problem of the same entity using multiple names (aliases).''(\cite{cite-niu})
\end{enumerate}
\end{quote}
On et al. \cite{cite-on} formally express their \textit{Name Disambiguation} problem as follows:
\begin{quote}
 ``Given two \textit{long} lists of author names, $X$ and $Y$, for each author name $x\ (\in X)$, find a set of author names, $y_1,y_2, ..., y_n\ (\in Y)$ such that both $x$ and $y_i\ (1 \leq i \leq n)$ are name variants of the same author.'' (\cite{cite-on})
\end{quote}
In contrast to the previous definitions Han et al. \cite{cite-han} define \textit{Name Dis\-ambi\-guation} like this:
\begin{quote}
``Name disambiguation can have several causes. Because of name variations, identical names, name misspellings or pseudonyms, two
types of name ambiguities in research papers and bibliographies (citations) can be observed. The first type is that an author has
multiple name labels. For example, the author `David S. Johnson' may appear in multiple publications under different name abbreviations such as `David Johnson', `D. Johnson', or `D. S. Johnson',
or a misspelled name such as `Davad Johnson'. The second type is that multiple authors may share the same name label. 
For example, 'D. Johnson' may refer to `David B. Johnson' from Rice University, `David S. Johnson' from AT\&T research lab, or `David
E. Johnson' from Utah University (assuming the authors still have these affiliations).''(\cite{cite-han})
\end{quote}
The citations above show that there are many expressions for Personal Name Matching (or sub-categories) which are not equally used by different authors. Niu et al. and On et al. restrict \textit{Name Disambiguation} to finding synonyms, Han et al. include homonyms in their definition.
Even more related expressions can be found in literature. As mentioned, we will use \textit{Personal Name Matching} in this work as Reuther uses it.
\\\\The main aspect of Personal Name Matching is handling synonyms and homonyms.
Trying to express the problems formally leads to the following description:
Let $X$ be a set of persons, especially characterized by their names, in a certain data set and $P$ a set of all existing persons.
We are also being given a function $Label(x_i) \rightarrow String$ and a relation $Entity = ((x_i,p_j) \mid p_j\ is\ the\ real\ person\ concealed\ by\ x_i)_{X \times P}$.
The actual problems can be described as
\begin{enumerate}[label=\arabic*)]
 \item $ \forall x_i,x_j \in X\ \exists\ (x_i,p_m),(x_j,p_n) \in Entity \mid Label(x_i) \not= Label(x_j) \wedge p_m = p_n$\label{enum-synonym}
 \item $ \forall x_i,x_j \in X\ \exists\ (x_i,p_m),(x_j,p_n) \in Entity \mid Label(x_i) = Label(x_j) \wedge p_m \not= p_n$\label{enum-homonym}
\end{enumerate}
with $1 \leq i \leq |X|$; $i \leq j \leq |X|$; $1 \leq m,n \leq |P|$.
\\\\Case \ref{enum-synonym} checks for each person $x_i$ from the person set $X$ whether another person $x_j$ from $X$ exists, so that their name labels are different ($Label(x_i) \not= Label(x_j)$) but the person is the same ($p_m = p_n$). So this case covers the synonym problem because the same person has several names here.
\\Case \ref{enum-homonym} checks for each person $x_i$ from the person set $X$ whether another person $x_j$ exists in $X$, so that their name labels are equal ($Label(x_i) = Label(x_j)$) but the persons behind the names differ ($p_m \not= p_n$). So this case covers the homonym problem because the same name is taken by several people.
\\The problem Personal Name Matching arises because such a relation $Entity$ usually does not exist and needs to be approximated as good as possible:
\[ Entity^* = ((x_i,p_j) \mid sim(x_i,p_j) \geq \theta \Rightarrow p_j\ is\ concealed\ by\ x_i)_{X \times P} \approx Entity \]
Thanks to appropriate similarity measurements and a matching threshold $\theta$, we can find such a relation $Entity^*$ which is approximately equivalent to the original relation $Entity$.
The main task in Personal Name Matching is finding a good similarity measure for the described problem. (\cite{cite-reuther}, p. 52)
\\\\Let us have a look at a vivid example.
\begin{figure}[t]
\begin{center}
  \caption{Synonyms and homonyms: Michael Keaton}\label{pic:keaton-douglas}
  \includegraphics[width=0.8\textwidth]{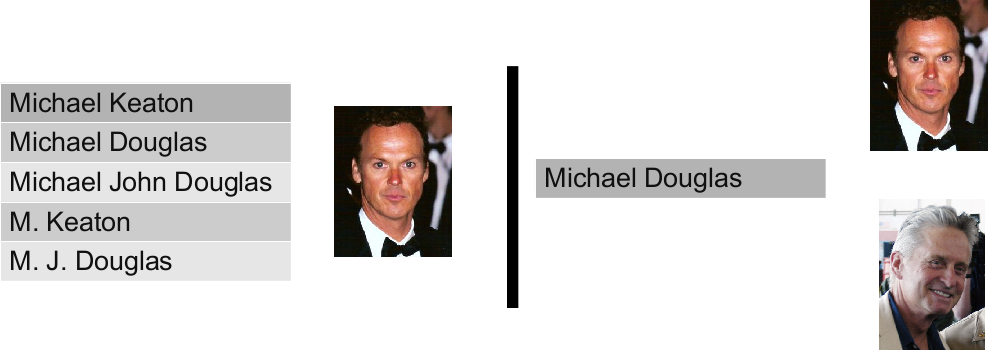}
  \caption*{Self-made creation, adapted from \cite{cite-reuther}, p.~51, photos from \url{http://en.wikipedia.org/}, photo of Michael Keaton by Georges Biard}
\end{center}
\end{figure}
The birth name of the famous actor Michael Keaton is Michael John Douglas\footnote{see entry about Michael Keaton at \url{http://www.imdb.com/name/nm0000474/bio}}. 
Keaton took a pseudonym because he could have been confused with the more famous actor Michael Douglas. Synonyms for Keaton are ``Michael Keaton'', ``Michael Douglas'',  ``Michael John Douglas'', ``Michael J. Douglas'', ``M. Keaton'' or ``M. J. Douglas''.
\enlargethispage{-1\baselineskip}
\\On the other hand, when we hear the name ``Michael Douglas'' we cannot be sure which famous actor is referred to, because Michael Douglas is a valid name for both of them. Figure \ref{pic:keaton-douglas} illustrates this Personal Name Matching problem with Michael Keaton.
\\\\The process of Personal Name Matching can be divided into the following steps (\cite{cite-reuther}, p. 56-87):
\begin{description}
  \item[Standardization] \hfill \\
  First the personal name data have to get standardized not to compare apples and oranges. Especially if the data source has a low data quality, this step is quite important. Otherwise, false or bad similarity values would be computed and assigned by the similarity function because the names are displayed by a different usage of upper and lower case letters or use national characters that are not used in the other name.
  \item[Blocking] \hfill \\
  We create blocks by assigning people to these blocks in this step. The goal is redu\-cing the number of name comparisons and thereby improving the runtime. Depending on whether we are looking for synonyms or homonyms, the blocking method differs. Known methods are \textit{Sorted Neighbourhood} \cite{cite-hernandez}, clustering \cite{cite-baxter}, phonetic hashing (\textit{Soundex} \cite{cite-russel}, \textit{Phonix} \cite{cite-gadd}, \textit{Metaphone} \cite{cite-philips}) or token based blocking \cite{cite-on}.
  \newpage
  \item[Analysis] \hfill \\
  This phase analyzes the names and contains the name comparisons. The task is finding possible synonyms and homonyms which will be evaluated in the next phase.
  \item[Decision Model] \hfill \\
  After the detection of possible synonyms and homonyms we have to evaluate the similarity of the names and make the decision to handle them as matches or not.
  \item[Performance Measurement] \hfill \\
  At last, we can evaluate the performance and correctness of the whole process. The purpose is finding out how reliable our data are and thinking about improvements to get an approximation to our optimal relation $Entity$ that is as close as possible.
\end{description}
Criteria for the evaluation of such a process are Precision and Recall (\cite{cite-baeza99}, p. 75-81; \cite{cite-reuther}, p. 83-85). Let $U$ be a set of items, $R$ be the set of relevant items (e.g. synonyms) with $R \subseteq U$ and $A$ be the answer of a request. In our scenario, the request is usually the question ``Is the item $u \in U$ a synonym, or accordingly $u \in R$?''.
Then we can define:
\[ Precision = \frac{|R \cap A|}{|A|} \]
\[ Recall = \frac{|R \cap A|}{|R|} \]
Precision testifies whether the reported synonyms during the Name Matching process are really synonyms, Recall allows us to say whether there are synonyms which have not been found.
\\We use a combination of the Jaccard Similarity Coefficient and Levenshtein Distance in our tool. Bilenko et al. \cite{cite-bilenko} explain these string matching methods isolated.
Given two word sets $S$ and $T$, the simple Jaccard Similarity Coefficient is:
\[ Jac(S,T) = \frac{|S \cap T|}{|S \cup T|} \]
\pagebreak[4]
\\
The Levenshtein Distance uses the operations replacement, insertion and deletion of a character and is defined by a matrix. Let $s \in S$ and $t \in T $ be words, $a = |s|$ and $b = |t|$ their lengths.
Then we can define:
\begin{multline}
   D_{0,0} = 0 \\
   D_{i,0} = i, \quad 1 \leq i \leq a \\
   D_{0,j} = j, \quad 1 \leq j \leq b \\
   D_{i,j} = min \left\{ 
  \begin{array}{l l}
    D_{i-1,j-1} & \quad \text{if } s_i = t_i \\
    D_{i-1,j-1} +1 & \quad \text{(replacement)} \\
    D_{i,j-1} +1 & \quad \text{(insertion)} \\
    D_{i-1,j} +1 & \quad \text{(deletion)} 
  \end{array} \right. 
  \mbox{with } 1 \leq i \leq a, 1 \leq j \leq b.
\end{multline}
We modify the Jaccard Similarity Coefficient in a way that it classifies two set items as intersected if their Levenshtein Distance is lower than a certain threshold.
\\\\In addition to the general Personal Name Matching, we must take the characteristics of Japanese names into account.
Particularly the usage of \textit{kanji} and several possibilities to transcribe a name make it hard to compare Japanese names.\footnote{see chapter \ref{writing-jap}}
For example, we cannot compare \textit{kanji} names from the IPSJ DL with the author names in DBLP. Even though \textit{kanji} are suited best for name comparison it does not work here because the standard encoding of names in DBLP is ``Latin-1'' which does not support \textit{kanji} natively.
\\A big problem for our work is revealed by looking at the given name \textit{Akiko} with its \textit{kanji} representation 章子.
As we can see in table \ref{tab:akiko} 章子\ has several possible readings besides \textit{Akiko} (left column) and \textit{Akiko} written in Latin characters does not determine a nonambiguous match in \textit{kanji} (right column).
\begin{table}[ht]
\begin{center}
\begin{tabular}{| c || c |}
  \hline
  \textbf{Possible readings of 章子} & \textbf{Possible writings for \textit{Akiko}} \\
  \hline
  \textit{Akiko} & 章子 \\
  \hline
  $Akirako$ & 明子 \\
  \hline
  $Atsuko$ & 晶子 \\ 
  \hline
  $Ayako$ & 愛季子 \\ 
  \hline
  $Sh\bar{o}ko$ & 亜記子 \\ 
  \hline
  $Takako$ & 暁子\\
  \hline
  & あき子 \\
  \hline
\end{tabular}
\caption{Problems with the given name \textit{Akiko}, sources: ENAMDICT; \cite{cite-eschbach}, p. 172-173}\label{tab:akiko}
\end{center}
\end{table}
\\The same problem applies to Japanese family names. Table \ref{tab:kojima} presents the problem with \textit{Kojima} as a family name example.
\begin{table}[ht]
\begin{center}
\begin{tabular}{| c || c |}
  \hline
  \textbf{Possible readings of 小島} & \textbf{Possible writings for \textit{Kojima}} \\
  \hline
  \textit{Kojima} & 小島 \\
  \hline
  \textit{Ojima} & 児島 \\
  \hline
  \textit{Koshima} & 児嶋 \\ 
  \hline
  \textit{Oshima} & 古島 \\ 
  \hline
  & 古嶋 \\ 
  \hline
\end{tabular}
\caption{Problems with the family name \textit{Kojima}, sources: ENAMDICT; \cite{cite-eschbach}, p. 223}\label{tab:kojima}
\end{center}
\end{table}
\chapter{Preparation of Japanese Papers for the Import Into the DBLP Data Set}\label{import}
\begin{center}
大事の前の小事 \quad Daiji no mae no sh\=oji
\\(Who wants to achieve big things must do the little things first.)
\\\textit{Japanese saying}
\end{center}

This chapter explains the approach to process and combine the various data sources so that we can import Japanese publications in the end. We will proceed step by step to make the ideas behind the solution as comprehensible as possible.
\section{General Approach}\label{approach}
First we will construct a table in a relational database containing information about Japanese names and their transcriptions by converting the ENAMDICT name dictionary.
Then we set up a data structure for Japanese names that handles the problem of assigning a given and a family name to a newly instantiated author during parsing the publications of \acs{IPSJ DL}.
At last, we will discuss the actual and titular integration of Japanese papers into the DBLP data set including an explanation that shows how to create a harvester for the OAI-PMH protocol.
\section{Converting an ENAMDICT File to a Relational Database}\label{enamdict-to-db}
The first step towards being able to handle Japanese names is distinguishing given and family name in the input text.
A relational database containing information about Japanese names and their transcriptions is useful for this task.
The database should contain names in \textit{kanji}, their transcriptions in \textit{hiragana} and Latin characters and the name type to have a good match with the data source ENAMDICT and to provide all necessary name information we need.
\\To fill the empty database, the ENAMDICT file needs to be analyzed and its data needs to be extracted. 
The entries usually have the form  
\begin{center}
KANJI [TRANSCRIPTION\footnote{the transcription in \textit{hiragana} is located at this place (we usually mean Latin characters when we talk about transcriptions, but not this time)}] /LATIN (TYPE)/\;.
\end{center}
We can take the following line as an example of an existing entry:
\begin{center}
森田\ [もりだ] /Morida (s)/
\end{center}
A parser should export the single entries.
First it saves the text between the slashes and searches for the type of the entry.
It must be assured that all person name types and no undesired or alleged types will be stored. Types can consist of the characters ``s'' (surname), ``g'' (given name), ``f'' (female name), ``m'' (male name), ``u'' (unclassified name), ``p'' (place name), ``h'' (full name of a particular person), ``pr'' (product name), ``co'' (company name) or ``st'' (station name).
But only the types ``s'', ``g'', ``f'' and ``m'' are important in this case because the parser should only store person names in the database. 
One exception are the unclassified names and they need to be stored too because they can also contain person names. 
Using unclassified names carelessly leads to problems, though. On the one hand it is useful if you find a match for the given name but not for the assumed family name. Then it helps to find an unclassified name matching the assumed family name. On the other hand some unclassified names in the ENAMDICT file decrease the data quality of the database. The entry
\begin{center}
スターウォーズ\ /(u) Star Wars (film)/
\end{center}
shows that there are undesired names like film titles in the category ``unclassified''. The example also reveals that there is no overall standard for an entry format.
Analyzing the file leads to following observations:
\begin{enumerate}[label=\alph*)]
 \item text in round brackets might be type or additional commentary (see entry example above)\label{enum-observ-a}
 \item when only \textit{hiragana} or \textit{katakana} are used instead of \textit{kanji} to display the Japanese name the transcription part is missing because it is not required (see entry example above)\label{enum-observ-b}
 \item the type information in brackets might actually consist of several type declarations, separated by commas\label{enum-observ-c}
 \item the type information might be placed before or after the transcription in Latin characters\label{enum-observ-d}
 \item one entry line might contain several possibilities to interpret the name, the example\label{enum-observ-e}
	\begin{center}
	イブ\ /(f) Eve/(u) Ib/Ibu (f)/(m) Yves/
	\end{center}
	clarifies this aspect
\end{enumerate}
We must consider these observations when we implement the parser.\\
To handle the problems in \ref{enum-observ-a} and \ref{enum-observ-c} we can filter the contents in round brackets. One possibility is using a regular expression like \texttt{(,|s|u|g|f|m|p|h|pr|co|st)$^*$} to filter all valid types. Regular expressions are powerful and popular tools for pattern matching. In our case we are looking for valid type expressions including commas to get rid of commentaries.
After eliminating commentaries we also want to get rid of unwanted types like place names. So we filter again and only process desired types this way.
To handle \ref{enum-observ-b} we just ignore missing transcriptions in square brackets.
Our parser also needs to be flexible enough to deal with observation \ref{enum-observ-d} which means that it must expect the type(s) at two possible places (before and after the transcription in Latin characters).
We can handle the last observation \ref{enum-observ-e} by using recursive function calls. We call the function that exports one entry with a modified parameter value within the function itself when there is more than one entry in the input line (noticeable by additional slashes).
\\Before parsing we need to change the original encoding of the ENAMDICT file from ``EUC-JP'' to ``UTF-8'' to make it compatible with our program.
\\During parsing a few inconsistencies in the syntax of the ENAMDICT file occurred:
\begin{itemize}
\item there were four times no slash in the end of the entry:
\begin{center}
甲子太郎\ [かしたろう] /Kashitarou (m)
\end{center}
\item there was once an unnecessary closing bracket without an opening bracket:
\begin{center}
近松秋江\ [ちかまつしゅうこう] /Chikamatsu Shuukou\textbf{\textcolor{red}{)}} (h)/
\end{center}
\item there was once a backslash where a square bracket was supposed to be put:
\begin{center}
キルギス共和国\ [キルギスきょうわこく\textbf{\textcolor{red}{\textbackslash}}\ /(p) Kyrgyz Republic/Kirghiz Republic/
\end{center}
\end{itemize}
Instead of constructing a workaround for these problems we should rather correct the only few inconsistencies manually.\footnote{The author decided to contact Jim Breen and told him about the found inconsistencies. Mr Breen corrected them in the ENAMDICT master file. So the inconsistencies should not exist anymore in the latest downloadable version.}
\section{A Data Structure for Japanese Names}\label{impl-structure-jnames}
We will construct a class which is responsible for handling Japanese names and representing them in a convenient way. Therefore, it must be able to save the name in \textit{kanji} and in at least one Latin transcription.
The transcription is necessary to compare found authors in IPSJ DL with authors in the DBLP. The \textit{kanji} name can be stored as additional author metadata in the DBLP later.
Our goal is a standardized representation of a Japanese person. So first we can construct a simple helper class for a single name containing given and family name as strings.
This class can be applied to both \textit{kanji} and Latin names. Our Japanese person usually has these two name representations.
\\When getting an input name from the IPSJ DL we try to determine the separation point and categorize the tokens into given and family names. The separation point can mostly be identified by white space or a comma between the words.
The categorization is done by including information from ENAMDICT. Thanks to ENAMDICT's classification into name types we can use this information to categorize our input name tokens into given and family names.
However, we have to cover some unusual cases too because IPSJ DL has no standardized way to provide names. So we get names in various formats.
For example, there are entries in which the family name follows the given name directly without any separation markers. Then we can try to take advantage of upper and lower case letters assuming that an uppercase letter means the beginning of a new name token. But we must also be aware of existing input names like ``KenjiTODA''. If we get a longer sequence of uppercase letters, this sequence is probably a family name. We can filter these names with a regular expression like \texttt{[A-Z][a-z]\{1,\}[A-Z]\{3,\}} (first character is an uppercase letter, followed by at least one lowercase letter, followed by at least three uppercase letters).
We also have to recognize abbreviated names and normalize Latin names.
\\Let us have a look at what we can observe about necessary transcription customizations. One peculiarity is that Japanese like to transcribe their names with an $h$ instead of a double vowel. An example is ``Hitoshi Gotoh''. The $h$ symbolizes the lengthening of a vowel and is a substitute for $o$ or $u$ in this case. To enable our class to find names like this in ENAMDICT, we have to replace the $h$'s lengthening a vowel by the vowel itself because ENAMDICT entries contain double vowels instead of $h$'s with this semantic function.
\\Another observation is ENAMDICT's usage of the Hepburn transcription system throughout the entire dictionary. So we have to convert the name to match the Hepburn system and to check a name via ENAMDICT. The needed character replacements for a conversion into the Hepburn system are shown in table \ref{tab:tohepburn} (see also figure \ref{pic:app-trans-diffs} in the appendix).
\begin{table}[ht]
\begin{center}
\begin{tabular}{| c | c |}
  \hline
  \textbf{Replaced character sequence} & \textbf{Replaced by} \\
  \hline
  tu & tsu \\
  \hline
  ti & chi \\
  \hline
  sya & sha \\
  \hline
  syo & sho \\
  \hline
  syu & shu \\
  \hline
  zya & ja \\
  \hline
  zyo & jo \\
  \hline
  zyu & ju \\
  \hline
  tya & cha \\
  \hline
  tyo & cho \\
  \hline
  tyu & chu \\
  \hline
  si & shi \\
  \hline
  hu & fu \\
  \hline
  zi & ji \\
  \hline
  jya & ja \\
  \hline
  jyo & jo \\
  \hline
  jyu & ju \\
  \hline
  l & r \\
  \hline
\end{tabular}
\caption{Conversion into the Hepburn transcription system}
\label{tab:tohepburn}
\end{center}
\end{table}
In addition to the replacements from table \ref{tab:tohepburn}, we must consider that names usually start with uppercase letters and replace ``Tu'', ``Ti'', ``Sya'' and so on by ``Tsu'', ``Chi'', ``Sha'', etc. as well.
\\The Japanese $n$ is sometimes transcribed as $m$. If $n$ is followed by $b$ or $p$, this $n$ is likely to be transcribed as $m$. The reason is a correlative modification in the pronunciation of $n$ in these cases. For example, the family name \textit{Kanbe} is often transcribed as \textit{Kambe} in the IPSJ DL data set.
\enlargethispage{-1\baselineskip}
\\Double vowels are sometimes completely dropped in some IPSJ DL author elements. While this might be okay for aesthetic reasons when transcribing the own name, it becomes a problem when we try to find a matching name in a dictionary like ENAMDICT. So we also have to check additional modified names. If there is a single vowel in the name, we must also check the same name whose vowel has become a double vowel. If several single vowels occur in a name, the number of names to be checked rapidly increases too. 
We have to pay special attention to the doubling of the vowel $o$ because $oo$ \uppercase{and} $ou$ are possible doublings for the single $o$. Doubling the vowel $e$ leads either to $ee$ or $ei$. All other double vowels are intuitive: $a$ becomes $aa$, $i$ becomes $ii$, $u$ becomes $uu$.
Taking ``Gotoh'' as an example we remove the $h$ first and check a list of names via ENAMDICT. The list of names consists of ``Goto'', ``Gooto'', ``Gouto'', ``Gotoo'', ``Gotou'', ``Gootoo'', ``Goutoo'', ``Gootou'' and ``Goutou''. We can remove ``Goto'', ``Gooto'' and ``Gouto'' from the list if we know that the $h$ (representing a double vowel) has been removed before.
\\If the input metadata contains a Latin and \textit{kanji} representation of the author's name, we will try to find a match for these. Names in \textit{kanji} usually do not have any se\-pa\-ration mark, so we must distinguish given and family name by taking advantage of the ENAMDICT dictionary and checking the possible name combinations.
Processing author names without \textit{kanji} representation is okay but a missing Latin representation becomes a problem when it comes to actually integrating the publication into the DBLP data set because all DBLP data are supposed to have a Latin representation.
The solution is a search for name candidates (we will discuss it more detailed in section \ref{create-bht}).
\\We cannot be sure that our name matching for Latin and \textit{kanji} names always succeeds. Therefore, we add some status information to our Japanese name to get a chance to evaluate the outcome of the program.
Possible status types are:
\begin{description}
 \item [ok] \hfill \\
 The status ``ok'' means that given and family name have successfully been found in the name dictionary and (if available) the \textit{kanji} names have successfully been assigned to their corresponding name in Latin characters.
 \item [undefined/Latin name missing] \hfill \\
 An undefined status usually means that the Latin name is missing. A missing Latin name leads to a never changed name status. In these cases, the name in \textit{kanji} usually exists anyway.
 \item [abbreviated] \hfill \\
 This is the status type for an abbreviated name like ``T. Nakamura''.
 \item [not found in name dictionary] \hfill \\
 If this status occurs, the Latin name could not be found in the name dictionary.
 \item [no kanji matching found] \hfill \\
 If a \textit{kanji} name has not been found in the name dictionary or could not be assigned to the Latin name, this status will occur.
 \item [bad data quality in source] \hfill \\
 As the name suggests, this status means that the data quality of the publication metadata source is most likely bad. Our tool can handle some of these cases well by normalizing the name.
 \item [possible name anomaly] \hfill \\
 We could have stumbled upon a name anomaly when we see this status type. During implementation this status was narrowed down to a possible name anomaly for abbreviated names.
 \item [name anomaly] \hfill \\
 This status indicates a critical name anomaly. This is the only case in which the tool cannot even give a recommendation for given and family name. The output is the full name of the input data for both given and family name.
\end{description}
In chapter \ref{p-name-matching} we discussed synonyms and homonyms. With the strategies from above we can deal with synonyms pretty well. Yet, homonyms cannot be recognized this way and are not covered at all by our tool.
\section{Import Into the DBLP Data Set}\label{import-into-dblp}
To be able to import the harvested data into the DBLP, we still need to make the existing publication data processable in an appropriate way for our program, construct a coauthor table for these data, compare publications from the Digital Library of the IPSJ with those available in the DBLP project and provide the new publication metadata for the DBLP adequately.
\subsection{Converting the DBLP XML File to a Relational Database}\label{dblp-to-sql}
It is important to convert the DBLP file $dblp.xml$ to a relational database to gain an easier and more efficient access to the data while running our program.
We are mainly interested in the basic publication metadata. So we will skip some non-publication records of the DBLP like $www$ elements\footnote{see also section \ref{create-bht} to get an example of a $www$ record}.
Our publication database table shall contain columns for an \acs{ID}, the authors, title, publication year, journal title, journal pages and the volume.
Whenever we come across the beginning of a publication type element ($article$, $inproceedings$, $proceedings$, $book$, $incollection$, $phdthesis$, $mastersthesis$) during parsing, we reinitialize the variables which store this metadata for the table columns.
When we encounter the according XML end tag of the publication we add an \acs{SQL} \textit{INSERT} command to a batch of commands. This batch is regularly executed after processing a certain amount of publications. The regular execution of batches allows a better performance than sending single \textit{INSERT} commands to the database server.
There are some recommendations in the DBLP FAQ \cite{cite-dblp-faq} for parsing the $dblp.xml$ file. We use the Apache Xerces\footnote{\url{http://xerces.apache.org/}} parser instead of the standard Java \acs{SAX} parser and need to increase the allocatable heap space for our parser.
\\While parsing the DBLP file we can construct a table with coauthor relationships along with the DBLP publication table. This coauthor table stores two author names and a publication \acs{ID}. The ID shows which publication has been written together by the authors and matches the ID in the DBLP publication table. 
New coauthor relationships will only be inserted if there are at least two authors mentioned in the metadata. If the metadata mentions more than two authors, every possible pair of authors will be inserted into the database.
\subsection{Implementation of an OAI-PMH Harvester}\label{impl-harvester}
As already explained in section \ref{ipsj-server}, we access the OAI-PMH repository by the repository name and the metadata format prefix to get a list of publication metadata entries.
The specification of \acs{OAI-PMH} 2.0 \cite{cite-oai-pmh} describes a possibility to retrieve a list of all metadata formats which a Data Provider has to offer. The \acs{HTTP} request
\begin{lstlisting}[]{http3}
http: //ipsj.ixsq.nii.ac.jp/ej/?action=repository_oaipmh
  &verb=ListMetadataFormats
\end{lstlisting}
informs us that there are two metadata formats called \textit{oai\_dc} and \textit{junii2}. \textit{oai\_dc}\footnote{\url{http://www.openarchives.org/OAI/2.0/oai_dc.xsd}, version from 19.12.2002} is the standard Dublin Core format all Data Providers provide, also traceable in the protocol specification. The ``Implementation Guidelines for the Open Archives Initiative Protocol for Metadata Harvesting'' \cite{cite-oai-guidelines} classify the metadata format \textit{oai\_dc} as mandatory. The name \textit{junii2}\footnote{\url{http://irdb.nii.ac.jp/oai/junii2.xsd}, state of 2012-10-10} suggests that it is a self-developed format of the National Institute of Informatics (in Tokyo).
Comparing these two in IPSJ DL, we notice that \textit{junii2} provides a more accurate description of the data, for example regarding additional XML attributes telling us whether the element value is English or Japanese. This additional information is helpful when we process the data in a later step and is missing in the \textit{oai\_dc} representation of the IPSJ server's data.
So we will take the metadata prefix \textit{junii2} as initial point for harvesting the server's metadata.
Figure \ref{pic:example-oai-pmh-junii2} shows an according metadata example (also compare figure \ref{pic:example-oai-pmh-dc}).
\begin{figure}[tb]
\begin{center}
  \caption{Part of publication metadata in \textit{junii2} format}\label{pic:example-oai-pmh-junii2}
  \includegraphics[scale=0.5]{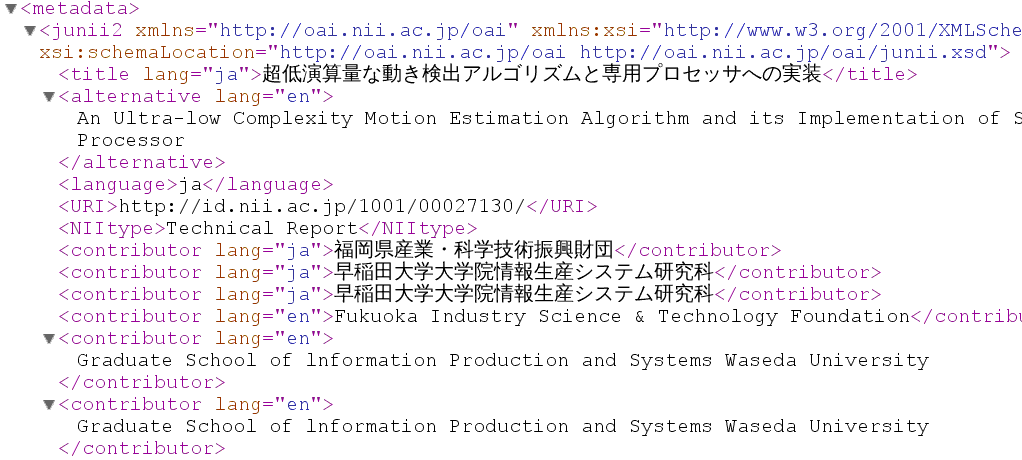}
\end{center}
\end{figure}
\\The harvesting includes the following steps:
\begin{itemize}
  \item we load the DBLP publication, coauthor relationship and the ENAMDICT data into the \ac{RAM}
  \item we access the IPSJ server to get publication metadata
  \item we parse the accessed XML metadata (concerning the thoughts from section \ref{impl-structure-jnames}) and store the needed publication data temporarily in the \acs{RAM}.
  \item we add the parsed publication to an SQL command batch to insert the metadata into a relational database (the batch is regularly executed)
  \item we create a \acs{BHT} file for the parsed publication\footnote{see section \ref{create-bht}}
  \item at the end we go into all directories with BHT files and concatenate them to one bigger BHT file
\end{itemize}
During the implementation and testing, some exceptional incidents occurred. We try to cover them besides the expected difficulties like Personal Name Matching and transcriptions.
For example, we get ``NobukazuYOSHIOKA'' as a full input name. Algorithm \ref{alg:nobukazu} shows a way to handle these unusual input data. Japanese sometimes write their family names in upper case letters to distinguish given and family name.
\begin{algorithm}[htb]
 \KwData{\begin{itemize}
	\itemsep0em
	\parskip0em
	\parsep0em
          \item $fullname$: full input name
          \item $names$: list of name representations for a Japanese person
          \item function split($text, regex$): searches for regular expression and splits text, 
	    \\ \quad splitted text does not contain text that matches the regular expression
          \item function normalize($name$): normalizes personal name
         \end{itemize}
        }
 \KwResult{new name for person found and added (given and family name separated)}
 \vspace{1 em}
  \If{$fullname$ matches regular expression $[A-Z][a-z]\{1,\}[A-Z]\{3,\}$}{
   $givenname \leftarrow$ split$(fullname, [A-Z]\{3,\})$\;
   $familyname \leftarrow$ split$(fullname, ([A-Z][a-z]\{1,\})$\;
   normalize$(familyname)$\;
   $status \leftarrow$ BAD\_DATA\_QUALITY\_IN\_SOURCE\;
   $names.$add(new PersonName$(givenname, familyname))$\;
  }
 \caption{Categorizing names like ``NobukazuYOSHIOKA''}
 \label{alg:nobukazu}
\end{algorithm}
\\Another observation during testing the program and checking the data is the following.
Searching the Japanese given name ``Shin'ichi'' in the DBLP we notice that there is no uniform way to store certain names in the database. We find ``Shin'ichi Aihara'' but also ``Shin-ichi Adachi'' along with other results indicating the same phenomenon. So we see the apostrophe and the hyphen are used equally as syllable separators (we discussed the syllable separation in chapter \ref{latin-chars}). 
Comparing the author ``Shinichi Horiden'' from the IPSJ data set and the one from the DBLP data set we can assume they are the same person because they have common coauthors (e.g. Kenji Taguchi and Kiyoshi Itoh) in both databases. The IPSJ data set tells us that the name written in \textit{kanji} is 本位田真一. We are interested in the part 真一\ (\textit{Shin'ichi}) because we get to know that the separator symbol is sometimes missing. The \textit{kanji} indicates the syllables $shi-n-i-chi$, especially focused on $n$ and $i$ instead of $ni$. We would expect an additional separator symbol for a clear (nonambiguous) transcription; but obviously, it has been dropped in this case. 
A separator symbol can also be found when some double vowels occur. For example, we find ``Toru Moto'oka'' (元岡達) instead of ``Toru Motooka''. This makes it easier to identify the reading of a single \textit{kanji} (元\ \textit{moto}, 岡\ \textit{oka}, 達\ \textit{toru}).
When a separator symbol is needed for a clear transcription, an apostrophe is used as separator symbol in ENAMDICT. While ENAMDICT always uses an apostrophe as separator symbol, DBLP and IPSJ DL use an apostrophe, a hyphen or the separator symbol is missing. We must consider these differences in the data sources for a successful import. For an easier name matching between names in the ENAMDICT and IPSJ DL data set we can add names containing an apostrophe once as they are and once without apostrophes to the relational database when we parse the ENAMDICT file to store person names in a relational database.
\\Our tool has a statistics class to get an overview over the parsed input data and the quality of the output data.
We will have a look at these statistics created after the harvest. There are 81597 records with publication metadata and 8562 records which are marked as $deleted$ in the parsed data. Figure \ref{pic:statistics-records} shows a visualization in pie chart form.
\begin{figure}[ht]
\begin{center}
  \caption{Statistics about the parsed records}\label{pic:statistics-records}
  \includegraphics[scale=0.5]{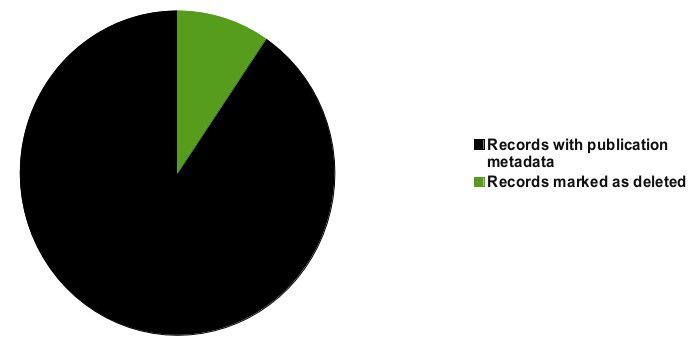}
  \caption*{Self-made creation}
\end{center}
\end{figure}
\pagebreak[4]
\\The publication types are declared as ``Technical Report'', ``Conference Paper'', ``Journal Article'', ``Departmental Bulletin Paper'' or ``Article'' (compare the table \ref{tab:pubtypes} and figure \ref{pic:statistics-pubtypes}).
\begin{table}[ht]
\begin{center}
\begin{tabular}{| l | r |}
  \hline
  \textbf{Publication types} & \textbf{Number of occurrences} \\
  \hline
  Technical Report & 51838 \\
  \hline
  Conference Paper & 6218 \\
  \hline
  Journal Article & 8401 \\
  \hline
  Departmental Bulletin Paper & 8694 \\
  \hline
  Article & 4295 \\
  \hline
\end{tabular}
\caption{The different publication types and the number of occurrences}
\label{tab:pubtypes}
\end{center}
\end{table}
\begin{figure}[!ht]
\begin{center}
  \caption{Statistics about the types of parsed publications}\label{pic:statistics-pubtypes}
  \includegraphics[scale=0.5]{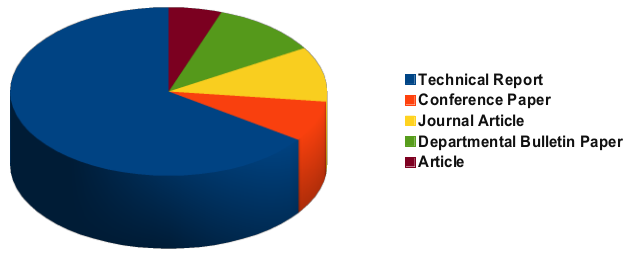}
  \caption*{Self-made creation}
\end{center}
\end{figure}
\\The statistics also reveal that 74971 publications are published in Japanese, only 4456 in English (compare the pie chart in figure \ref{pic:statistics-language}).
\begin{figure}[ht]
\begin{center}
  \caption{Statistics about the language of parsed publications}\label{pic:statistics-language}
  \includegraphics[scale=0.5]{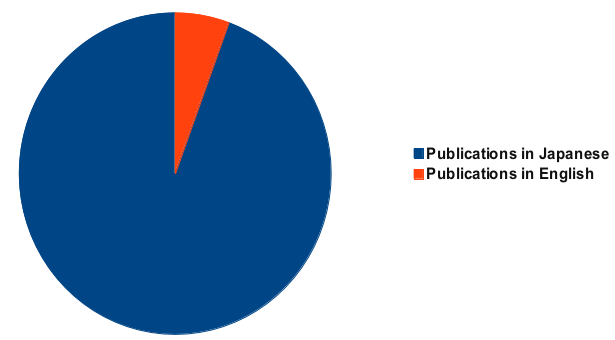}
  \caption*{Self-made creation}
\end{center}
\end{figure}
\\Our tool detects 1325 publications which are already included in DBLP. A publication is considered found in both databases if the title is the same and at least one author is the same.
\\The most interesting statistics for our work are these about the evaluation of the quality of author name assignments (compare the bar chart in figure \ref{pic:statistics-status}):
\begin{table}[!ht]
\begin{center}
\begin{tabular}{| l | r | r |}
  \hline
  \textbf{Status} & \textbf{Number of occurrences} & \textbf{Percentage} \\
  \hline
  ok & 180221 & 78.0\% \\
  \hline
  undefined & 25012 & 10.8\% \\
  \hline
  abbreviated & 132 & 0.1\% \\
  \hline
  not found in name dictionary & 9073 & 3.9\% \\
  \hline
  no kanji matching found & 14827 & 6.4\% \\
  \hline
  bad data quality in source & 1203 & 0.5\% \\
  \hline
  possible name anomaly & 393 & 0.2\% \\
  \hline
  name anomaly & 301 & 0.1\% \\
  \hline
\end{tabular}
\end{center}
\end{table}
\begin{figure}[!ht]
\begin{center}
  \caption{Statistics about the evaluation of the quality of author name assignments}\label{pic:statistics-status}
  \includegraphics[scale=0.5]{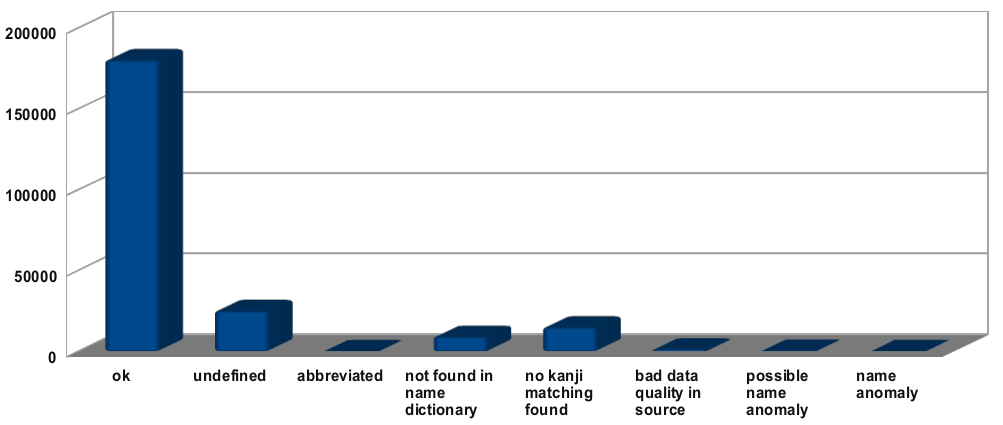}
  \caption*{Self-made creation}
\end{center}
\end{figure}
\\Fortunately, 180221 of 231162 author names could be matched successfully. There are many reasons for the remaining uncovered cases. 
9073 Latin names could not be found in the name dictionary ENAMDICT and 14827 name matchings between the names' Latin and \textit{kanji} representations did not succeed. 
These names might be missing at all in the dictionary, delivered in a very unusual format that the tool does not cover, or might not be Japanese or human names at all.
Of course, Japanese computer scientists sometimes also cooperate with foreign colleagues but our tool expects Japanese names and is optimized for them. 
Both IPSJ DL and ENAMDICT provide \textit{katakana} representations for some Western names. However, \textit{katakana} representations for Western names are irrelevant for projects like DBLP.
But for instance, Chinese names in Chinese characters are relevant. Understandably, our tool does not support any special Personal Name Matching for Chinese names yet because our work is focused on Japanese names.
The tool does not take account of the unclassified\footnote{compare section \ref{enamdict-to-db}} names of ENAMDICT by default. We can increase the general success rate of the Name Matching process by enabling the inclusion of unclassified names in the configuration file but the quality of the Name Matching process will decrease because the correct differentiation between given and family name cannot be guaranteed anymore.
An unclassified name may substitute a given or a family name.
\\There are 1203 entries that were qualified as ``bad data quality in publication metadata source''. They might be handled alright but they are particularly marked to indicate that these cases should also be reviewed manually before any import action is performed.
\\The numbers of abbreviated names, possible name anomalies and name anomalies are very low. While processing author names which will be later qualified as ``possible name anomaly'', the tool cannot decide whether the assignment has been correct or the name is an anomaly. ``Name anomalies'' are critical anomalies that could not be categorized into any other status.
\\There could be a few uncovered flaws, for example \acs{HTML} or \LaTeX\ code in titles. We must be aware of those when we do the actual import into the DBLP data set.
\subsection{Creation of BHT Files for Japanese Papers}\label{create-bht}
We will discuss the creation of \acs{BHT} files and important extensions for the BHT format that fit the requirements of Japanese papers well, based on our knowledge from section \ref{dblp}.
As mentioned, the header dictates ISO-8859-1 (Latin-1) as encoding of the file $dblp.xml$. Ley's work \cite{cite-ley-dblp} reveals that we can use \acs{XML}/\acs{HTML} entities to solve this problem. Authors have person records in the DBLP providing additional information. For example, we can find the following entry for Atsuyuki Morishima (森嶋厚行) in the XML file:
\begin{lstlisting}[language=XML]{}
<www mdate="2008-02-20" key="homepages/m/AtsuyukiMorishima">
  <author>Atsuyuki Morishima</author>
  <title>Home Page</title>
  <url>http://www.kc.tsukuba.ac.jp/~mori/index.html</url>
  <note>&#x68EE;&#x5D8B;&#x539A;&#x884C;</note>
</www>
\end{lstlisting}
We must extend the BHT format to fulfill the requirements and add extra metadata for authors, title and relevant process information.
The author talked to members of the DBLP team personally and got the permission to extend the original BHT format to enable us to adapt the format to Japanese papers.
Our additions are well formed XML elements. We must substitute all non-\acs{ASCII}\footnote{ASCII is a encoding scheme for characters of the alphabet and consists of 128 characters.} characters by escape characters (XML entities) to ensure the compatibility for DBLP.
The additional elements are:
\begin{itemize}
 \item Every author that has a \textit{kanji} representation in its metadata gets an \textit{originalname} element:
  \begin{lstlisting}[language=XML]{}
    <originalname latin="Shinsuke Mori">&#x68EE;,&#x4FE1;&#x4ECB;
      </originalname>
  \end{lstlisting}
  If available, the Latin representation is added as an attribute $latin$ to avoid confusion on assigning the extra information to the right author later on.
  The element content has a fixed structure. The family name comes first, followed by a comma and the given name.
 \item Every author gets a status information that evaluates the author name assignment. It is displayed by a \textit{status} element:
    \begin{lstlisting}[language=XML]{}
      <status name="Shinsuke Mori">ok</status>
    \end{lstlisting}
    The connected author is added as an attribute $name$.
 \item If there is no Latin representation of the name of an author, we will add Latin name candidates to the BHT file:
    \begin{lstlisting}[language=XML]{}
      <namecandidates kanji="&#x83C5;&#x8C37;&#x6B63;&#x5F18;">Shougu Sugatani, Seihiro Sugatani, Tadahiro Sugatani, Masahiro Sugatani, Shougu Suganoya, Seihiro Suganoya, Tadahiro Suganoya, Masahiro Suganoya, Shougu Sugaya, Seihiro Sugaya, Tadahiro Sugaya, Masahiro Sugaya, Shougu Sugetani, Seihiro Sugetani, Tadahiro Sugetani, Masahiro Sugetani, Shougu Sugenoya, Seihiro Sugenoya, Tadahiro Sugenoya, Masahiro Sugenoya</namecandidates>
    \end{lstlisting}
    The connected \textit{kanji} representation is added as an attribute \textit{kanji} in the \textit{namecandidates} element.
    We seek the \textit{kanji} in ENAMDICT and output all possible name combinations in a comma separated list.
 \item If the original language of the title is Japanese, we will add this title to the BHT file:
    \begin{lstlisting}[language=XML]{}
      <originaltitle lang="ja" type="Journal Article">&#x70B9;&#x4E88;&#x6E2C;&#x306B;&#x3088;&#x308B;&#x81EA;&#x52D5;&#x5358;&#x8A9E;&#x5206;&#x5272;</originaltitle>
    \end{lstlisting}
    The XML element \textit{originaltitle} has the attributes \textit{lang} (for the paper language) and \textit{type} (for the publication type).
 \item The tool searches the authors in DBLP and tries to find additional common coauthors in DBLP. If at least two of the main authors of the paper also worked with a certain other person (that is retrieved from DBLP), this person is added to the comma separated list. The Personal Name Matching of author names uses a combination of Levenshtein Distance and Jaccard Similarity Coefficient here.
    \begin{lstlisting}[language=XML]{}
      <commoncoauthors>Masato Mimura</commoncoauthors>
    \end{lstlisting}
 \item If the tool finds the paper in DBLP, we also add the DBLP key. Records, such as elements with publication metadata, have a unique key in DBLP.
    \begin{lstlisting}[language=XML]{}
      <dblpkey>conf/iscas/HiratsukaGI06</dblpkey>
    \end{lstlisting}
\end{itemize}
An example of a BHT file in \ac{SPF} can be found in the appendix in section \ref{app-bht} (also compare with the original BHT format in section \ref{app-dblp-bht}).
After we have finished parsing all Japanese papers, we concatenate the BHT files in \ac{SPF} that belong together to one bigger BHT file $all.bht$. Publications, respectively BHT files, that belong together are recognizable by the directory structure. If they belong together, they will be in the same directory.
We must simply go through the BHT root directory recursively.
\chapter{Conclusion and Future Work}\label{conclusion}
\begin{center}
``Creativity is seeing what everyone else sees, \\but then thinking a new thought that has never been \\thought before and expressing it somehow.'' \\(Neil deGrasse Tyson)
\end{center}

The integration of Japanese papers into the \acs{DBLP} data set has revealed some major problems. 
The nonambiguous representation of Japanese names (and paper titles, etc.) is done by \textit{kanji} while DBLP's standard encoding is Latin-1 and Japanese characters are only optionally added to the publications' metadata.
This leads to the need of transcribing the Japanese names which in turn also evokes new problems because there is not \textit{the} transcription but rather a lot of transcription possibilities.
\\In addition to that, we must ensure a certain data quality even if one data source sometimes lacks this quality.
Due to name matching with a name dictionary, format checking and conversions (if necessary), we can actually correct some flaws or at least assimilate the data into our project.
\\The problem of synonyms is dealt with by transcription manipulations, homonyms could not be addressed in this work.
Reuther (\cite{cite-reuther}, p. 159-164) describes an idea to \mbox{handle} homonyms.
We could extend our tool by a Coauthor Index\footnote{see section \ref{dblp}} as in DBLP for the publications of the IPSJ DL.
The idea is based on the assumption that scientists often publish their papers with the same people as coauthors. If the coauthors match a certain coauthor group, the author is considered the same.
\enlargethispage{-1\baselineskip}
If the author's coauthors are not members of the expected coauthor groups, the author could be a different person than we expected and we might have a homonym here.
\\\\The developed tool is usable and provides among relational databases customized Bibliography Hypertext (\acs{BHT}) files as output data.
Customizations were necessary to optimize the BHT files for Japanese papers and additional important metadata information.
Desired but missing metadata like contributors or a short description of the content of a paper can be added without much effort because the relational database already contains these data, only the source code of \textit{Kankoukanyuu} (our tool) needs to be extended by a few lines.
\\Though having been created with care regarding correct and well-formed output data, it is not re\-com\-mended to import the newly created BHT files unchecked.
The DBLP team should check the files not to compromise the data quality of DBLP.
There might still be undesired format anomalies in the BHT files. The DBLP team also needs to adapt their import system to the extended BHT format developed in this work for the actual import into DBLP.
\\Titles might be in uppercase letters. This could be improved but we have to pay attention because a primitive solution will not work well. For example, we have to be aware of the popular usage of acronyms in computer science. So some words in uppercase letters can be correct.
\\Our tool is optimized for the Digital Library of the IPSJ and their OAI-PMH metadata prefix \textit{junii2}. It can easily be adapted to support the similar and commonly used metadata prefix \textit{oai\_dc}. So the tool would be able to handle other publication metadata sources that support OAI-PMH.
\\The algorithm\footnote{algorithm mentioned in section \ref{impl-harvester}} for detecting common papers in DBLP and IPSJ DL may be modified to achieve an even better comparison between the databases and detect more common papers.
\\It would be useful to include a Chinese name dictionary in the future and extend the name search of our tool to cover Chinese names as well.
\enlargethispage{-1\baselineskip}
\\One improvement in the future could be storing the most common names (for example, the 100 most common given and family names) in a separate data structure in the \acs{RAM}.
This way we can improve the runtime by often skipping the search in the huge name data.
\\We can still increase the success rate of the Name Matching process too. One way is swapping \textit{kanji}. A typical Japanese name has two \textit{kanji} for the given name and two \textit{kanji} for the family name. The family name shall precede the given name. However, this principle could be violated by the publication source. If the Name Matching is not successful, we may swap the first two for the last two characters and try to find a match again.
\\A second advancement is the additional support of a special Latin character set that is used by Japanese. For instance, we can find the name ``Ｋａｉ'' instead of ``Kai'' in the metadata of IPSJ DL.
They look very similar and both represent simple Latin letters but their character codes are different. So programs handle them differently.
A simple (but yet unimplemented) substitution function can cover these rare and unusual cases.
\\Another possibility to take advantage of this work is extracting the author names in \textit{kanji} from the relational database. So the DBLP team can insert author metadata for already existing authors in DBLP.
\\\\We can also have a look at what phases of the Personal Name Matching process\footnote{see chapter \ref{p-name-matching}} have been implemented in this work and to which degree.
There are actually different types of Personal Name Matching included in our tool:
\begin{itemize}
 \item we try to match transcribed Japanese names with their original \textit{kanji} representation
 \item we try to distinguish given and family name (name in \textit{kanji} not necessarily needed)
 \item if a paper is also found in DBLP, we will try to look for common coauthors
\end{itemize}
The ``Standardization'' is accomplished by a normalization of the Latin input names at the beginning of the process. \textit{Kanji} input names get trimmed by removing all whitespace.
We do not have a ``Blocking'' phase as it is proposed by Reuther \cite{cite-reuther}.
When searching a match between transcribed Japanese names with their original \textit{kanji} representation we even go a contrary way and increase the number of comparisons by adding reasonable other transcriptions to the matching process.
Due to efficient data structures and a comparatively small amount of Japanese papers (less than 100000), our tool has an acceptable runtime (the retrieval of the publication metadata from the IPSJ server takes much longer than processing it).
In addition, the search for common coauthors will only be done if the author exists in DBLP.
The phases ``Analysis'' and ``Decision Model'' are entangled in our tool. If we find a match between a (normalized or modified) input name and a name in the name dictionary, we will immediately consider them a successful match and continue parsing the metadata.
When we try to find coauthors in DBLP, we take advantage of the combined Jaccard Levenshtein Distance as explained in chapter \ref{p-name-matching}.
\\Instead of checking the complete output data in the ``Performance Measurement'' phase, we could only take control samples while implementing, debugging, testing and improving our program.
A broad manual check of approximately 90000 publications is not possible within the scope of a diploma thesis.
The control samples had the expected and desired content but we cannot guarantee the correctness of the output.
Under the assumption that ENAMDICT's entries are correct, the predicted Precision\footnote{see definition in chapter \ref{p-name-matching}} should be about $1.0$ because the tool probably does not produce many false positives.
But we cannot say anything about the Recall because ENAMDICT does not cover all names that occur in IPSJ DL.
All exceptions resulting from the limits of a name dictionary and a bad data quality are supposed to be handled by the status for author name assignments (described in section \ref{impl-harvester}).
This gives us the chance to manually handle the noted exceptions afterwards.
\\\\All in all, this work is a first approach for an integration of Japanese papers into the DBLP data set and provides a not yet perfect but usable tool for this task. Some major obstacles are overcome.
\newpage
\phantomsection
\renewcommand\bibname{References}
\addcontentsline{toc}{chapter}{References}
\bibliography{bibliography}

\newpage
\appendix
\chapter{The Tool}\label{tool}
\section{About the Tool}\label{about-tool}
The developed tool that is also part of this project is named \textit{Kankoukanyuu} (刊行加入). \textit{Kankou} means publication, \textit{kanyuu} means admission. The whole name indicates the ability to import publications. The tool also allows the assimilation of imported publications, of course. The usable functionalities are:
\begin{itemize}
 \item Parsing the DBLP file $dplb.xml$ and converting it to a MySQL database
 \item Converting an ENAMDICT name dictionary file to a MySQL database
 \item Harvesting the IPSJ server, processing the publication metadata and storing it in a MySQL database
 \item Making the harvested publications ready for an import into the DBLP data set by making BHT files
\end{itemize}

\section{Usage}\label{usage}
The tool has been developed and tested on a Linux system with Intel Core 2 Quad and 8 \acs{GB} \acs{RAM} in the local computer pool.
It has to be executed by command line like this:
\begin{lstlisting}[]{exec}
java -Xmx5400M -jar kankoukanyuu.jar
\end{lstlisting}
The parameter \textit{-Xmx5400M} allows our program to allocate more than 5 GB RAM and store all necessary data in the RAM for an unproblematic execution.
\\\\Possible command line arguments are:
\begin{description}
  \item[-{-}parse-dblp, -d] \hfill \\
  Parse dplb.xml and fill database tables
  \item[-{-}enamdict, -e] \hfill \\
  Convert ENAMDICT dictionary file to a relational database
  \item[-{-}harvest, -h] \hfill \\
  Harvest the IPSJ server, fill OAI-PMH data into databases and create BHT files (in \acs{SPF}) - \textit{requires DBLP and ENAMDICT database tables from steps above}
  \item[-{-}concatenate-bht, -b] \hfill \\
  Concatenate BHT files in Single Publication Format to one bigger file (file \textit{all.bht} will be created in every folder with BHT files) - \textit{requires BHT files in \acs{SPF} from step above}
  \item[-{-}all, -a] \hfill \\
  Do all of the above
  \item[-{-}help, -help] \hfill \\
  Show help text about usage of the tool
\end{description}
The configuration file $config.ini$ allows us to change following parameters:
\begin{itemize}
  \item Database related parameters (in $[db]$ section): \acs{URL} ($url$), database name ($db$), user name ($user$) and password ($password$)
  \item ENAMDICT related parameter (in $[enamdict]$ section): location of ENAMDICT file ($file$)
  \item ENAMDICT database related parameters (in $[japnamesdb]$ section): database table name ($table$), decision whether to use unclassified names ($useunclassifiednames$)
  \item DBLP related parameter (in $[dblp]$ section): location of $dblp.xml$ ($xmlfile$)
  \item DBLP database related parameters (in $[dblpdb]$ section): database table name for publications ($dblptable$), database table name for coauthor relationships (\textit{authorscounttable})
  \item OAI-PMH database (contains output after harvest and parsing process) related parameters (in $[oaidb]$ section): publication table ($publicationtable$), authors table ($authorstable$), titles table ($titlestable$), contributors table ($contributorstable$), descriptions table ($descriptionstable$)
  \item Harvester related parameters (in $[harvester]$ section): location for storing the harvest ($filespath$), start ID for harvester ($minid$), end ID for harvester ($maxid$), decision whether to use record lists ($uselistrecords$)
  \item BHT export related parameters (in $[bhtexport]$ section): location for BHT output files ($path$), decision whether to compute and show common coauthors (\textit{showcommoncoauthors})
  \item Log related parameter (in $[log]$ section): location of log files ($path$)
\end{itemize}
A configuration example can be found in the appendix section \ref{app-config}.
\\\\The system must support the Japanese language (meaning Japanese characters) to ensure a successful run. 
\\\textit{Kankoukanyuu} does not use any Linux-only commands but has not been tested on Microsoft Windows yet.
\section{Used Technologies}\label{used_technologies}
The tool itself has been written in Java, using the Open\acs{JDK} 6. The handling of databases is done by My\acs{SQL} 5 and \ac{JDBC} is used to provide MySQL functionalities within Java.
\\External libraries are the Apache Xerces\footnote{\url{http://xerces.apache.org/}} parser and the MySQL Connector/J\footnote{\url{http://dev.mysql.com/downloads/connector/j/}}.
The Fat Jar Eclipse Plug-In\footnote{\url{http://fjep.sourceforge.net/}} is used to deploy the complete project into one executable Java \acs{JAR}\footnote{Archive which includes compiled Java code, project metadata, libraries and possibly more} file. 
The execution of \textit{Kankoukanyuu} becomes more user-friendly this way because external libraries are already included and class paths for external libraries does not need to be specified anymore.
\section{Runtime}\label{runtime}
Measurement indicates the following approximated runtimes of \textit{Kankoukanyuu}:
\begin{table}[ht]
\begin{center}
\begin{tabular}{| l | l |}
  \hline
  \textbf{Activity} & \textbf{Runtime} \\
  \hline
  Parsing DBLP & 1 hour, 8 minutes \\
  \hline
  Conversion of ENAMDICT & 1 minute \\
  \hline
  Harvest by record lists, filling of & 5 hours, 11 minutes \\database tables and creation of BHT files &  \\
  \hline
  Concatenation of BHT files & 1 to 6 minutes \\
  \hline
\end{tabular}
\end{center}
\end{table}
\\We can make some observations. During the harvest, only ca. 30 minutes were spent on processing the harvested data, the rest is needed to retrieve the data from the Japanese server.
Depending on whether the local file system or network file system was used, the runtime for the concatenation differs immensely.
\chapter{Additional Material}
\newpage
\section{Differences Between the Transcription Systems}\label{app-trans-diffs}
\begin{figure}[h!]
\begin{center}
  \caption{Differences between the transcription systems}\label{pic:app-trans-diffs}
  \includegraphics[scale=0.25]{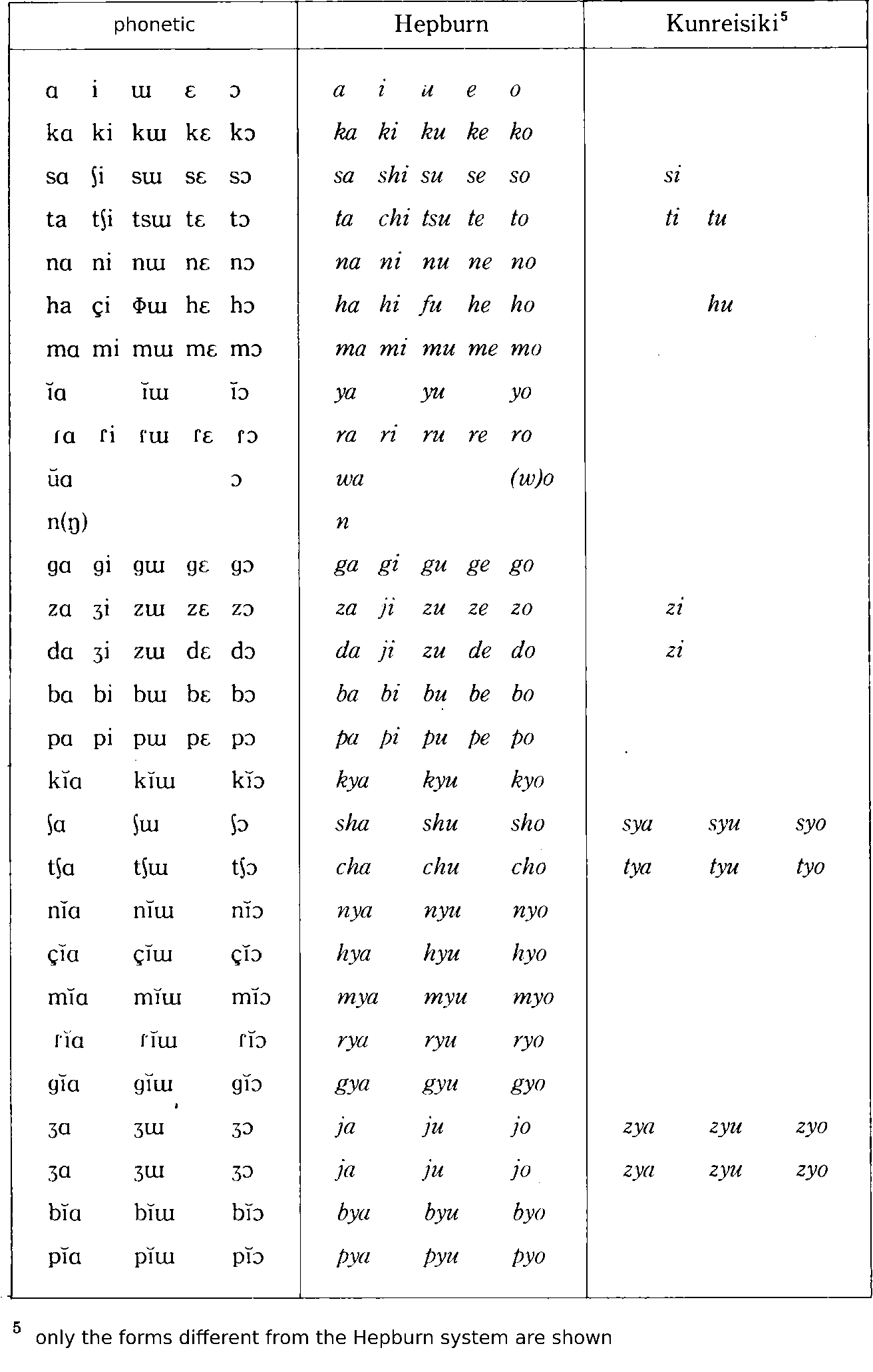}
  \caption*{Source: \cite{cite-saito}, p. 26, translated}
\end{center}
\end{figure}
\newpage
\section{OAI-PMH Example: \textit{junii2} Element}\label{app-oai-xml}
\begin{figure}[h!]
\begin{center}
  \caption{OAI-PMH example: \textit{junii2} XML element from IPSJ server}\label{pic:oai-xml}
  \includegraphics[width=\textwidth]{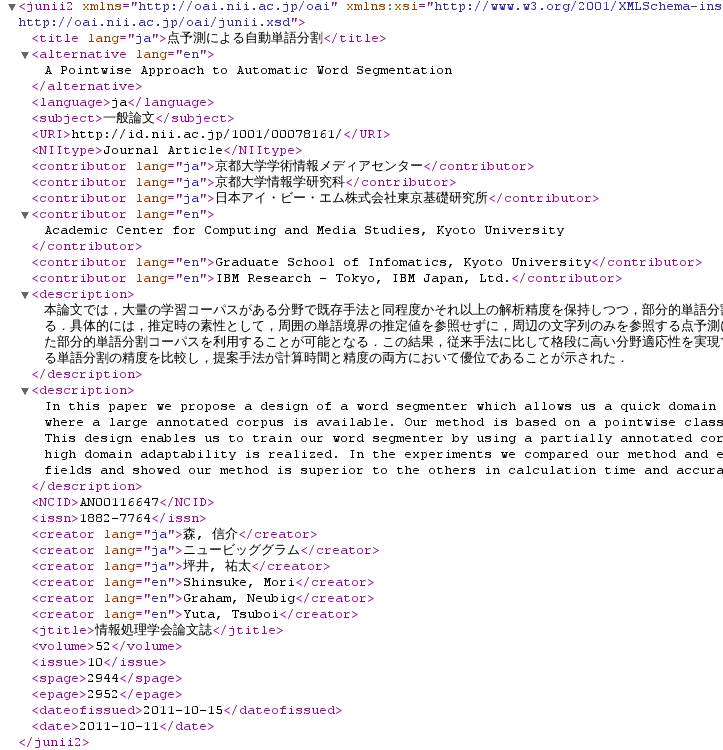}
\end{center}
\end{figure}
\newpage
\section{BHT Example Proposed By DBLP}\label{app-dblp-bht}
\begin{lstlisting}[language=XML]{}
Computer Languages, Systems &amp; Structures  (journals/cl)

<h2>Volume 34, Numbers 2-3, July-October 2008</h2>
Best Papers 2006 International Smalltalk Conference
<ul>
<li>Wolfgang De Meuter:
Preface.
45
<ee>http://dx.doi.org/10.1016/j.cl.2007.07.001</ee>
<li>David R&ouml;thlisberger, Marcus Denker, &Eacute;ric Tanter:
Unanticipated partial behavioral reflection: Adapting applications at runtime.
46-65
<ee>http://dx.doi.org/10.1016/j.cl.2007.05.001</ee>
<li>Johan Brichau, Andy Kellens, Kris Gybels, Kim Mens, Robert Hirschfeld, Theo D'Hondt:
Application-specific models and pointcuts using a logic metalanguage.
66-82
<ee>http://dx.doi.org/10.1016/j.cl.2007.05.004</ee>
<li>Alexandre Bergel, St&eacute;phane Ducasse, Oscar Nierstrasz, Roel Wuyts:
Stateful traits and their formalization.
83-108
<ee>http://dx.doi.org/10.1016/j.cl.2007.05.003</ee>
<li>Alexandre Bergel, St&eacute;phane Ducasse, Colin Putney, Roel Wuyts:
Creating sophisticated development tools with OmniBrowser.
109-129
<ee>http://dx.doi.org/10.1016/j.cl.2007.05.005</ee>
<li>Luc Fabresse, Christophe Dony, Marianne Huchard:
Foundations of a simple and unified component-oriented language.
130-149
<ee>http://dx.doi.org/10.1016/j.cl.2007.05.002</ee>
</ul>
\end{lstlisting}
This is a BHT example proposed by the DBLP team in the DBLP FAQ \cite{cite-dblp-faq}.\footnote{\url{http://dblp.uni-trier.de/db/about/faqex1.txt}, accessed at 2013-01-15}
\newpage
\section{BHT Example File Created By \textit{Kankoukanyuu}}\label{app-bht}
\begin{lstlisting}[language=XML]{}
<h2>Volume 52, Number 10, October 2011</h2>
<ul>
<li>Shinsuke Mori, Graham Neubig, Yuuta Tsuboi:
A Pointwise Approach to Automatic Word Segmentation.
2944-2952
<ee>http://id.nii.ac.jp/1001/00078161/</ee>
<originalname latin="Shinsuke Mori">&#x68EE;,&#x4FE1;&#x4ECB;</originalname>
<status name="Shinsuke Mori">ok</status>
<originalname latin="Graham Neubig">&#x30CB;&#x30E5;&#x30FC;&#x30D3;&#x30C3;&#x30B0;&#x30B0;&#x30E9;&#x30E0;,</originalname>
<status name="Graham Neubig">no kanji matching found</status>
<originalname latin="Yuuta Tsuboi">&#x576A;&#x4E95;,&#x7950;&#x592A;</originalname>
<status name="Yuuta Tsuboi">ok</status>
<originaltitle lang="ja" type="Journal Article">&#x70B9;&#x4E88;&#x6E2C;&#x306B;&#x3088;&#x308B;&#x81EA;&#x52D5;&#x5358;&#x8A9E;&#x5206;&#x5272;</originaltitle>
<commoncoauthors>Masato Mimura</commoncoauthors>
</ul>
\end{lstlisting}
This is an output example of a BHT file in Single Publication Format (before the concatenation step), created by our tool.
\newpage
\section{Excerpt From \textit{dblp.xml}}\label{app-dblp.xml}
\begin{lstlisting}[language=XML]{dblp.xml}
<?xml version="1.0" encoding="ISO-8859-1"?>
<!DOCTYPE dblp SYSTEM "dblp.dtd">
<dblp>
<article mdate="2002-01-03" key="persons/Codd71a">
<author>E. F. Codd</author>
<title>Further Normalization of the Data Base Relational Model.</title>
<journal>IBM Research Report, San Jose, California</journal>
<volume>RJ909</volume>
<month>August</month>
<year>1971</year>
<cdrom>ibmTR/rj909.pdf</cdrom>
<ee>db/labs/ibm/RJ909.html</ee>
</article>

<article mdate="2002-01-03" key="persons/Hall74">
<author>Patrick A. V. Hall</author>
<title>Common Subexpression Identification in General Algebraic Systems.</title>
<journal>Technical Rep. UKSC 0060, IBM United Kingdom Scientific Centre</journal>
<month>November</month>
<year>1974</year>
</article>

<article mdate="2002-01-03" key="persons/Tresch96">
<author>Markus Tresch</author>
<title>Principles of Distributed Object Database Languages.</title>
<journal>technical Report 248, ETH Z&uuml;rich, Dept. of Computer Science</journal>
<month>July</month>
<year>1996</year>
</article>
...
\end{lstlisting}
\newpage
\section{Configuration File of Our Tool}\label{app-config}
\begin{lstlisting}[]{config.ini}
[db]
url=myserver
db=mydbname
user=myusername
password=mypassword
[japnamesdb]
table=japnames
useunclassifiednames=false
[dblpdb]
authorscounttable=dblpauthors
dblptable=dblp
[oaidb]
publicationtable=oai_publications
authorstable=oai_authors
titlestable=oai_titles
contributorstable=oai_contributors
descriptionstable=oai_descriptions
[enamdict]
file=./enamdict
[harvester]
filespath=./files-harvester
minid=1
maxid=100000
uselistrecords=true
[dblp]
xmlfile=/dblp/dblp.xml
[bhtexport]
path=./bht
showcommoncoauthors=true
[log]
path=./log
\end{lstlisting}

\end{Japanese}
\end{CJK}

\end{document}